\documentclass[10pt,twocolumn,letterpaper]{article}

\usepackage{wacv}              

\usepackage{graphicx}
\usepackage{amsmath}
\usepackage{amssymb}
\usepackage{booktabs}
\usepackage[accsupp]{axessibility}  

\usepackage[pagebackref,breaklinks,colorlinks]{hyperref}

\usepackage[capitalize]{cleveref}
\crefname{section}{Sec.}{Secs.}
\Crefname{section}{Section}{Sections}
\Crefname{table}{Table}{Tables}
\crefname{table}{Tab.}{Tabs.}

\usepackage{makecell}
\usepackage{multirow}

\newcommand\yj[1]{\textcolor{black}{#1}}

\newcommand{\JH}[1]{{\color{black}{#1}}}
\newcommand\by[1]{\textcolor{black}{#1}} 
\newcommand\cam[1]{\textcolor{black}{#1}}

\def\wacvPaperID{1399} 
\def\confName{WACV}
\def\confYear{2024}

\begin{document}

\title{EResFD: Rediscovery of the Effectiveness of Standard Convolution for Lightweight Face Detection}


\author{Joonhyun Jeong$^\textrm{1,2}$ \quad
Beomyoung Kim$^\textrm{1,2}$ \quad
Joonsang Yu$^\textrm{1,3}$ \quad
YoungJoon Yoo$^\textrm{1}$\\
$^\textrm{1}$NAVER Cloud, ImageVision \quad
$^\textrm{2}$KAIST \quad
$^\textrm{3}$NAVER AI Lab \\
{\tt\small \{joonhyun.jeong,beomyoung.kim,joonsang.yu,youngjoon.yoo\}@navercorp.com}
}

\maketitle

\begin{abstract}

This paper \cam{analyzes} the design choices of face detection architecture that improve efficiency of computation cost and accuracy. Specifically, we re-examine the effectiveness of the standard convolutional block as a lightweight backbone architecture for face detection. 
Unlike the current tendency of lightweight architecture design, which heavily utilizes depthwise separable convolution layers, we show that heavily channel-pruned standard convolution layer\cam{s} can achieve better accuracy and inference speed when using a similar parameter size.
This observation is supported by the analyses concerning the characteristics of the target data domain, face\cam{s}.
Based on our observation, we propose to employ ResNet with a highly reduced channel, which surprisingly allows high efficiency compared to other mobile-friendly networks (e.g., MobileNetV1, V2, V3).
From the extensive experiments, we show that the proposed backbone can replace that of the state-of-the-art face detector with a faster inference speed. 
Also, we further propose a new feature aggregation method to maximize the detection performance.
Our proposed detector EResFD obtained 80.4\% mAP on WIDER FACE Hard subset which only takes 37.7 ms for VGA image inference on CPU. Code is available at \url{https://github.com/clovaai/EResFD}.

\end{abstract}

\section{Introduction}
\label{sec:intro}

\by{
\yj{Face detection research \cam{has demonstrated} significant performance improvement after \cam{the} advent of recent deep neural network based} general object detection approaches such as one-stage detector ($e.g.,$ SSD~\cite{ssd}, YOLO~\cite{redmon2016you(yolo)}, RetinaNet~\cite{lin2017focal(retinanet)}, EfficientDet~\cite{tan2020efficientdet(efficientdet)}) and two-stage detector ($e.g.,$ Faster R-CNN~\cite{ren2015faster}, FPN~\cite{fpn}, Mask R-CNN~\cite{he2017mask(mrcnn)}, Cascade R-CNN~\cite{cai2018cascade(cascadercnn)}).
For \cam{applicability} in a real-world scenario, real-time face detection has attracted more attention, and recent face detectors commonly adopt the one-stage approach that is simpler and more efficient than the two-stage approach.
}



\begin{figure}
\centering
	\includegraphics[width=0.95\columnwidth]{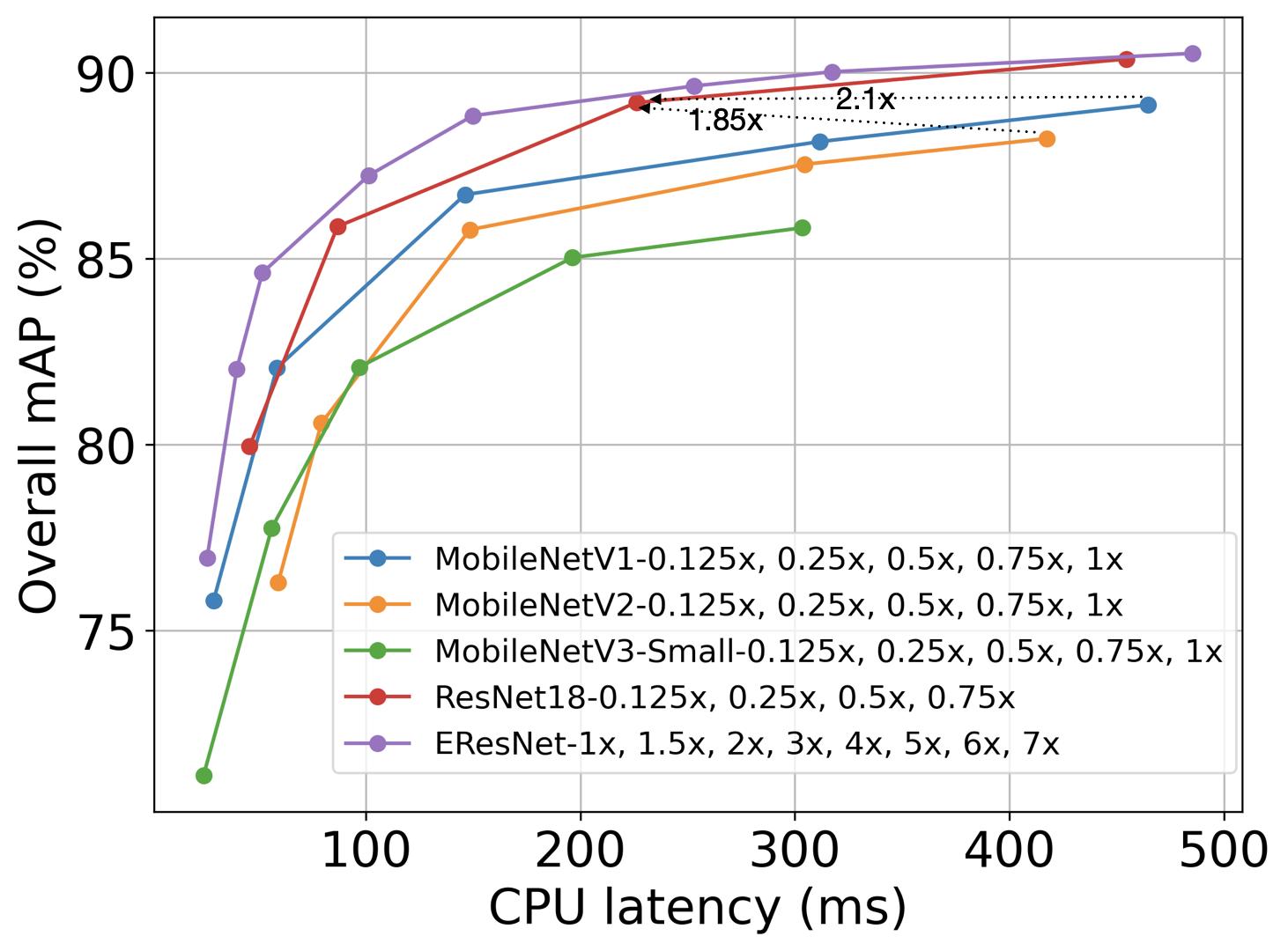}
	\caption{Latency-accuracy Pareto curve. We investigate the latency trend of \cam{depthwise separable convolution-based backbones (MobileNets) and standard convolution-based backbones (ResNets)} according to adjusting width multiplier \JH{on RetinaFace\cite{deng2019retinaface} framework}. ResNet shows much lower latency compared with MobileNets family even though its mAP is higher than others. 
	\yj{One step forward to the observation, we propose a modified ResNet backbone for face detection task, abbreviated as EResNet, which reports \cam{far more accurate and faster performance than the others.}} 
	}
	\label{fig:latency}
\end{figure}


\by{
Recent studies for real-time face detection methods frequently use the lightweight model consisting of depth-wise separable convolution, which is used in MobileNet~\cite{howard2017mobilenets(mobilenet)} and ShuffleNet~\cite{zhang2018shufflenet(sufflenet)}.
Specifically, \yj{recent lightweight face detectors}, such as RetinaFace~\cite{deng2019retinaface}, SCRFD~\cite{guo2021sample}, and CRFace~\cite{vesdapunt2021crface}, employ MobileNetV1 architecture~\cite{howard2017mobilenets} as the backbone network and reduce the number of channels in depthwise separable convolution layers by adjusting the width multiplier.
Following the paradigm of residual block~\cite{resnet}, BlazeFace~\cite{bazarevsky2019blazeface} proposed BlazeBlock that consists of depthwise separable convolution layers with skip connection, achieving a stronger performance.
In practice, adopting the depthwise separable convolution is a reasonable choice to save the number of floating point operations (FLOPs), which is one of the important measurement\cam{s} for the real-time application.
\yj{Summing up the} following common practice, most real-time face detectors utilize the depthwise separable convolution layers in their model by default~\cite{deng2019retinaface, guo2021sample, wu2023yunet}.
}

\by{
In this paper, we rethink the common belief for the depthwise separable convolution layer and found out that the standard convolution with reducing the number of channels can achieve a better trade-off between latency and detection performance than depthwise separable convolution.
Here, we use ResNet18~\cite{resnet} as our baseline backbone network for the standard convolution and compare with the depthwise separable convolution-based backbone networks (MobileNetV1~\cite{howard2017mobilenets(mobilenet)}, MobileNetV2~\cite{sandler2018mobilenetv2}, and MobileNetV3~\cite{howard2019searching}).
\figurename~\ref{fig:latency} shows the latency and average of mean average precision (mAP) scores on WIDER FACE~\cite{widerface} Easy, Medium, and Hard subsets. ResNet18 demands much higher latency than MobileNet when the width multiplier is not applied ($i.e., 1$x).
However, ResNet18 becomes much faster than MobileNet with higher mAP when reducing the number of channels using the width multiplier.
Note that ResNet18-0.5x denotes that width multiplier 0.5 is applied, and it is 2.1 times faster than MobileNetV1-1.0x and 1.85 times faster than MobieNetV2-1.0x even though its mAP is higher than others.
}

\yj{Based on the observation}, we propose EResFD, which is a ResNet-based real-time face detector.
\JH{We \yj{firstly propose a slimmed version of} ResNet architecture, namely EResNet, by redesigning the new stem layer, and changing the block configuration.}
Those methods can effectively reduce the inference latency and achieve higher detection accuracy compared with ResNet18.
\yj{Secondly}, we also propose the new feature map enhancement modules; Separated Feature Pyramid Network (SepFPN) and Cascade Context Prediction Module (CCPM).
SepFPN aggregates information from high-level and low-level feature maps separately, and CCPM further effectively captures diverse receptive fields by employing a cascade design.
\JH{Equipped with these architectural designs, our EResFD achieved 3.1\% higher mAP on WiderFace Hard subset compared to the state-of-the-arts lightweight face detectors such as FaceBoxes\cite{zhang2019faceboxes}. }

We summarize the main contributions as follows:
\begin{itemize}
\item We propose a ResNet-based extremely lightweight backbone architecture, which is much faster than \cam{the baselines} on the CPU \cam{devices}, achieving state-of-the-art detection performance. 
\item We analyze the behavior of both standard convolution and depthwise separable convolution, and we found that the standard convolution is much faster than the depthwise separable convolution under extremely lightweight \cam{parameter constraints}.
\item We propose a channel dimension preserving strategy to reduce the latency, fitting the number of layers in each layer group to recover the performance degeneration.
\item We propose a latency-aware feature enhance module, SepFPN, and CCPM. These enhance modules improve the detection performance on all (large, medium, small) face scales, with much faster speed compared to previous enhance modules. 
\end{itemize}

\section{Related Works}



\paragraph{Face Detectors}
Recent face detectors~\cite{s3fd,ssh,pyramidbox,dsfd,guo2021sample,deng2019retinaface,zhu2020tinaface,cao2021emface,li2019pyramidbox++,zhang2019accurate,zhu2020progressface,najibi2019fa,liu2020bfbox,yoo2019extd,liu2020hambox,ramos2020swiftface,bazarevsky2019blazeface,hoang2020deface,zhang2020asfd,song2020kpnet, qi2023fast} achieved impressive  performance enhancement.
These face detectors \cam{inherit} the architectural improvement of the general object detectors such as SSD~\cite{ssd} and RetinaNet~\cite{lin2017focal(retinanet)} or two-stage detectors such as Faster R-CNN~\cite{ren2015faster}.
The improvement in face detection enabled to detect faces with various densities and scales.
To detect dense and small-scale faces, current state-of-the-art group of detectors\cite{pyramidbox,li2019pyramidbox++,dsfd,guo2021sample,deng2019retinaface,zhu2020tinaface,liu2020hambox,li2018dsfd} mostly employ large-scaled classification networks with custom-designed upsampling blocks.
Besides the ResNet families~\cite{resnet}, prototypical choice, various attempts including those from architecture  search~\cite{zhang2020asfd} have been applied. 
PyramidBox series~\cite{pyramidbox,li2019pyramidbox++} and DSFD~\cite{li2018dsfd} suggested own upsampling blocks to improve the expressiveness of the features for dealing with finer faces. 
RetinaFace~\cite{deng2019retinaface}, currently the dominant one, infers five-point keypoint landmarks of the faces: eyes, nose, mouth, in addition to the detection box, similar to MTCNN~\cite{zhang2016joint}. 
However, the large memory size requirements of these face detectors critically hinder their applicability on edge devices. Here, we target on reducing the weight parameters of the backbone network to increase the usability of the face detectors.

\begin{figure*}[t]
    \centering
    \includegraphics[width=0.75\textwidth]{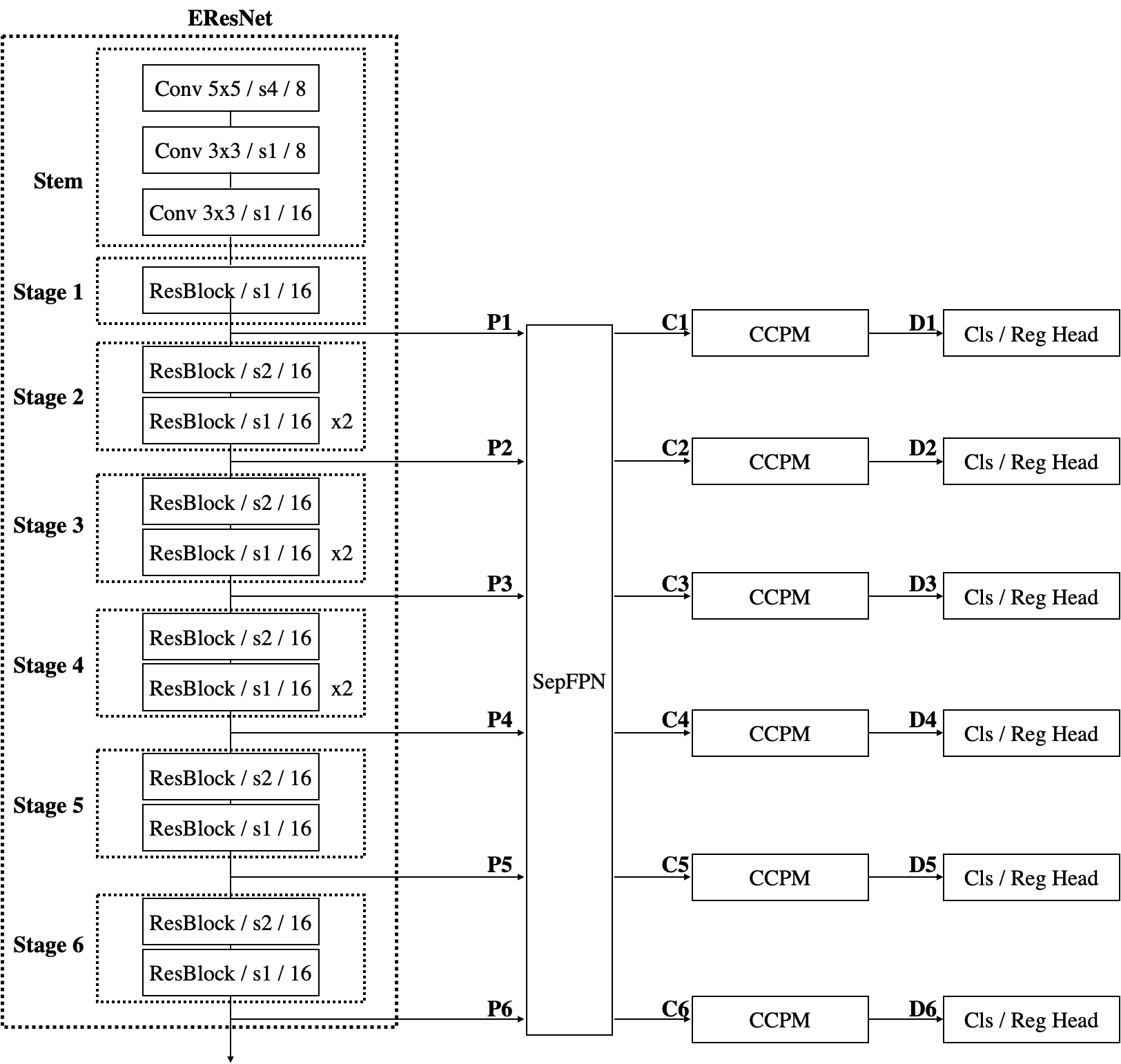}
    \caption{Entire architecture of EResFD. 
    The proposed architecture consists of EResNet with 31 weighted \cam{backbone} layers, 
    Separated Feature Pyramid Network (SepFPN), and Cascade Context Prediction Module (CCPM).
    ResBlock denotes the basic residual block, which was proposed in \cite{resnet}. 
    The first ResBlock of each stage has stride of 2, and every ResBlock has the same number of output channels as 16 in the case of EResFD-1x.
    For the classification and regression head, a single 1x1 convolution layer is used.}
    \label{fig:eresfd_arch}
    \vspace{-3mm}
\end{figure*}
\paragraph{Lightweight Face Detectors}
To run the above-mentioned face detectors on mobile or CPU devices, some of the detectors provide their lighter version, mostly substituting their backbones to lighter classification networks utilizing depthwise convolution~\cite{howard2017mobilenets}.
After advent of the pioneering works from MobileNetV1~\cite{howard2017mobilenets} and V2~\cite{sandler2018mobilenetv2} utilizing depthwise separable convolution and inverted bottleneck block, more refinements~\cite{wu2018shift,howard2019searching,han2020ghostnet,tan2019efficientnet,tan2021efficientnetv2} on the architectures have brought the performance enhancements. These architectures show the competitive ImageNet~\cite{deng2009imagenet} classification accuracy to larger classification models and also for the transferred tasks like object detection~\cite{ren2015faster,ssd} and segmentation~\cite{chen2018encoder}.
Following the improvement of the lightweight backbone networks,  RetinaFace~\cite{deng2019retinaface}, SCRFD~\cite{guo2021sample}, \cam{and YuNet~\cite{wu2023yunet}} use channel-width pruned version of MobileNet~\cite{howard2017mobilenets}. BlazeFace~\cite{bazarevsky2019blazeface} and MCUNetV2~\cite{lin2021mcunetv2}  proposed new variants of MobileNet targeting on mobile GPU and CPU environment, and EXTD~\cite{yoo2019extd} recursively uses the inverted bottleneck block of the MobileNet for further slimming the network size.
Besides the overall tendency of using depthwise separable convolution-based backbones,  KPNet~\cite{song2020kpnet} proposes its own backbone network consisting of standard convolutional network, but its size is still large for edge devices, about 1 million parameters, and focuses on sparse and large scaled faces. 
In this paper, we rediscover the efficiency of standard convolution layers, which can cover faces with various scales and densities, under extremely lightweight model size and minimal inference time.




\section{EResFD}
As seen in Figure~\ref{fig:latency}, \yj{ResNet with standard convolution achieves both faster inference time and higher detection performance \cam{compared to the widely used backbone, MobileNets~\cite{howard2017mobilenets, sandler2018mobilenetv2, howard2019searching} which heavily uses depthwise separable convolution layers.}
From this observation}, we revisit the ResNet architecture. 
\figurename~\ref{fig:eresfd_arch} illustrates the proposed face detection architecture, named as efficient-ResNet (EResNet) based Face Detector, EResFD.
It consists of two main parts; modified ResNet backbone architecture and \yj{ newly proposed feature enhancement modules.}
We modify several parts of ResNet to reduce the latency while preserving the detection performance \yj{based on empirical analysis on the network,} and we also propose both the new feature pyramid module and context prediction module, which are called Separated Feature Pyramid Network (SepFPN) and Cascade Context Prediction Module (CCPM), respectively. 
Both modules improve the detection performance, \yj{and also} show comparable or even faster latency compared to previous state-of-the-art CPU detectors \cite{faceboxes_improved, jin2019learning}.

\subsection{Rethinking ResNet Architecture}

\subsubsection{Convolutional Layer Analysis}
\begin{figure*}[t!]
    \centering
    \begin{subfigure}[t]{0.48\textwidth}
        \includegraphics[width=0.99\columnwidth]{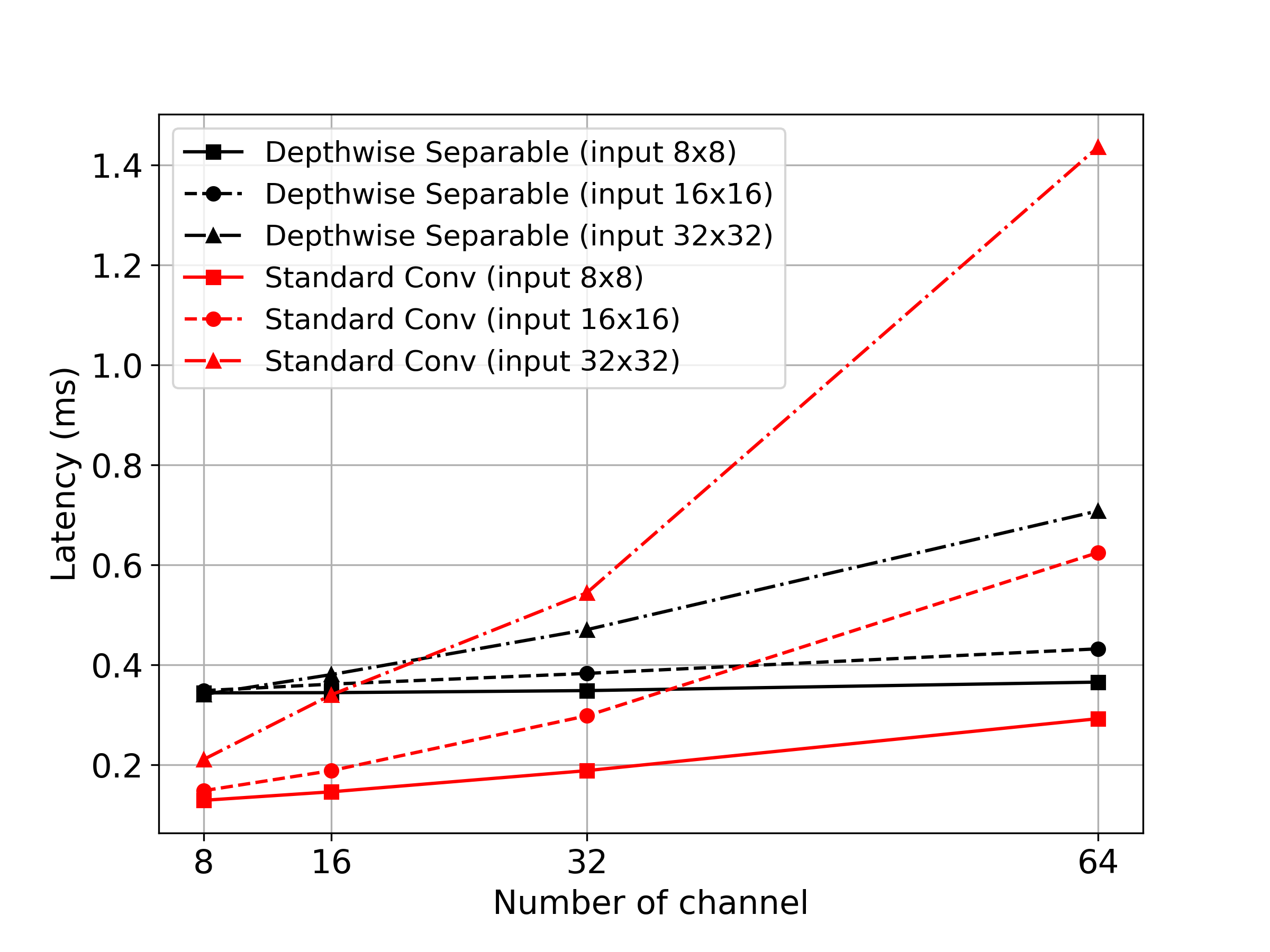}
        \caption{standard convolution vs depthwise separable convolution}
        \label{fig:conv_comparison}
    \end{subfigure}
    \begin{subfigure}[t]{0.48\textwidth}
        \includegraphics[width=0.99\columnwidth]{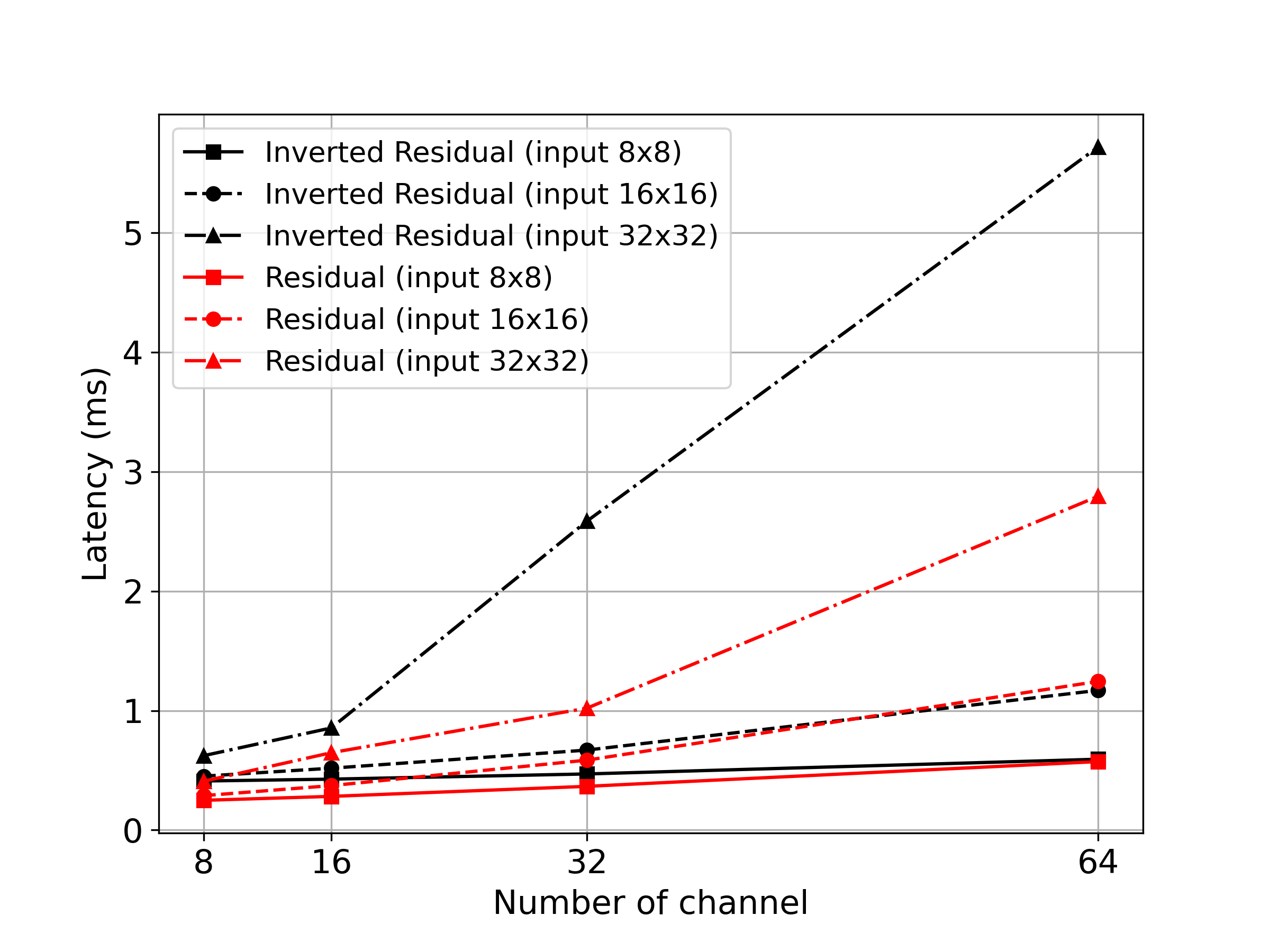}
        \caption{residual block (ResNet) vs inverted bottleneck block (MobileNetV2)}
        \label{fig:block_comparison}
    \end{subfigure}


    \caption{Illustration of latency comparison with varying channel size: (a) standard and depthwise separable convolution, (b) residual block (ResNet) and inverted bottleneck block (MobileNetV2).}
    \label{fig:latency_comp}
\vskip -0.1in
\end{figure*}

Depthwise separable convolution is introduced to reduce the multiplication and accumulation \yj{cost of the convolution}, which occupy most of the computation time during the inference.
Table~\ref{tab:dwc} shows the comparison of computational cost between the standard and depthwise separable convolution for each stage.
Depthwise separable convolution has much smaller FLOPs, \yj{
and hence} it can significantly reduce the computational cost.
However, previous work~\cite{bello2021revisiting} claimed that FLOPs is not always matched with actual latency.
Latency can be bounded by memory access and hardware accelerator, $i.e.,$ CPU or GPU, so the target hardware characteristic should be considered \yj{for the network design.}


\begin{table}[t]
    \footnotesize
    \setlength\extrarowheight{2pt}
    \centering
    \caption{Comparison of computational cost between standard and depthwise separable convolution. We calculate the FLOPs count for three kinds of setting, and each value indicates the multiply-add count for single layer.}
    \label{tab:dwc}
    \begin{tabular}{c|c|cc}
    \toprule
    Type & Standard & \multicolumn{2}{c}{Depthwise Separable} \\[0.5ex]
    \hline
    Operation & 3x3 Conv & Depthwise Conv & Pointwise Conv \\ [0.5ex]
    \hline
    \makecell{FLOPs \\ (H,W=16, C=16)} & $1.18$M & $0.07$M & $0.13$M \\ [0.5ex]
    \makecell{FLOPs \\ (H,W=16, C=32)} & $4.71$M & $0.15$M & $0.52$M \\ [0.5ex]
    \makecell{FLOPs \\ (H,W=16, C=64)} & $18.87$M & $0.29$M & $2.10$M \\ [0.5ex]
    \bottomrule
    \end{tabular}
    \vspace{-2mm}
\end{table}
To check the relationship between FLOPs and latency, we investigate the behavior of both standard convolution and depthwise separable convolution on CPU.
We measured the latency of both convolutional layers on CPU, and \figurename~\ref{fig:conv_comparison} shows the comparison result. \JH{Considering faster inference with small-sized input image ($e.g.,$ less than 320x), we tested with input sizes $8{\times}8$, $16{\times}16$, and $32{\times}32$.}
As input size increases, the latency of standard convolution is steeply increasing, but depthwise separable convolution shows a small amount of latency growth.
However, standard convolution achieves smaller latency than \yj{depthwise separable convolution on the extremely lightweight condition.} For all the input sizes, standard convolution is faster than depthwise separable convolution when its channel dimension is equal to or smaller than 16 as shown in \figurename~\ref{fig:conv_comparison}.
\yj{Since ResNet and MobileNets each consist of standard convolution and depthwise separable convolution, respectively, we can conclude} that ResNet has a chance to become faster than MobilNets when we extremely reduce the channel size.

Furthermore, we also analyze the block-level behavior of each convolutional layer.
We use residual block and inverted residual block, which consist of standard and depthwise separable convolution, respectively.
The residual block consists of two standard $3{\times}3$ convolution.
MobileNetV2 has an inverted residual block, which includes one depthwise convolution and two pointwise convolution.
The inverted residual block commonly expands the number of channels for the depthwise convolution, which is called the expansion ratio.
In MobileNetV2, the expansion ratio is set to 6 for most inverted residual blocks \cite{sandler2018mobilenetv2}, which is reported to preserve the classification ability of the block compared to standard convolution counterpart~\cite{han2022learning}.
Here, we use the equivalent expansion ratio for the latency comparison.
\figurename~\ref{fig:block_comparison} shows the block-level latency, and we found that residual block is much faster than the inverted residual block in most cases.
The residual block has $9.43$M FLOPs and the inverted residual block has $7.18$M FLOPs when the input size is $16{\times}16$ and the number of channels is 32.
Even though residual block has more multiply-add operations, its latency is faster than inverted residual block.

The latency trend of each layer and block shows that standard convolution has a chance to surpass the depthwise separable convolution \cam{in terms of} the latency.
This trend is also the same on network-level analysis as we mentioned in Section~\ref{sec:intro}.
Therefore, we propose an efficient backbone originating from the ResNet.

\subsubsection{Stem Layer Modification}

\begin{table}[]
    \small
    \centering
    \caption{Latency breakdown of ResNet18-0.25x model. Stem denotes 7x7 convolution layer followed by maxpool layer, which reduces spatial size by 4 times. For Stage 1 $\sim$ 4, strides of output feature map are set to 4 $\sim$ 32.}
    \label{tab:latency_breakdown}
    \begin{tabular}{c|cc}
    \toprule
    Component & Latency (ms) & Ratio (\%) \\
    \hline
    Stem & 24.1 & 44 \\
    Stage 1 & 10.5 & 19 \\
    Stage 2 & 7.5 & 13 \\ 
    Stage 3 & 6.6 & 12 \\ 
    Stage 4 & 6.5 & 12\\ 
    \hline
    Total & 55.2 & 100 \\
    \bottomrule
    
    \end{tabular}
\end{table}
We \cam{further thoroughly analyzed the ResNet architecture and observed} that the stem layer occupies a large amount of entire latency.
Table \ref{tab:latency_breakdown} shows the latency breakdown of ResNet18 with width multiplier $0.25$, and it shows that almost half of the total latency is originated from stem layers.
The backbone network is highly lightened by applying the small width multiplier, so the proportion of the stem layer becomes larger.
Moreover, ResNet stem layer consists of $7{\times}7$ convolution with stride of 2, so it requires a large amount of computation compared to others.
The number of computations (FLOPs) is proportional to the square of kernel size ($K^2)$ and reciprocal-square of stride ($1/S^2)$.
If kernel size is reduced to $5$, its FLOPs becomes about $50$\% of $7{\times}7$ convolution, and FLOPs further decrease to about $13$\% when its stride of 4 is applied simultaneously.

To reduce the latency of stem layers, we first change stride to 4 for the convolutional layer, which is already adopted in the previous work \cite{faceboxes_improved}.
We also reduce the kernel size from $7$ to $5$, but it can hurt the detection performance because it is directly related to the receptive field.
To alleviate this problem, we introduce two additional convolutional layers right after $5{\times}5$ convolution.
\cam{Owing to} those convolutional layers, the receptive field size becomes larger than the original stem layer, but its computation complexity is still much lower than the original.
\begin{table}[t]
    \small
    \centering
    \caption{Latency of stem layers on ResNet18-0.25x model. Ratio denotes the portion of stem latency compared to the overall network latency. 
    }
    \smallskip
    \noindent
    \begin{tabular}{c|cc}
    \toprule
    Stem & ResNet & EResNet \\
    \hline
    Stem FLOPs & $180.6$ M & $11.5$ M \\
    Stem Latency (Ratio) & 24.1ms (44\%) & 4.5ms (13\%) \\ \bottomrule
    \end{tabular}
    \label{table:stemlayer_modification_latency}
\end{table}
Table \ref{table:stemlayer_modification_latency} shows comparison results on the stem layer.
By adopting a smaller kernel size and bigger stride, EResNet stem layer has much smaller FLOPs, and also achieves much shorter latency compared to the original ResNet stem layer.


\subsubsection{Architecture Reconfiguration for Face}

\begin{table}[t]
    \small
    \centering
    \caption{Latency breakdown of ResNet18 models where channels are doubled or preserved for stage 2,3,4. Width multiplier is set to be 0.25 for ResNet-preserved model to keep number of output channels as 16 for all the stages.}
    \vspace{-4mm}
    \smallskip
    \noindent
    \begin{tabular}{cc|cc}
    \noalign{\smallskip}\noalign{\smallskip}
    \toprule
    \multicolumn{2}{c|}{Model} & ResNet & ResNet-Preserved  \\
    \hline 
    \multirow{2}{*}{Stage 2} & Latency (ms) & 7.5 & 4.2 \\
    & FLOPs (M) & 157.3 & 45.5 \\
    \hline
    \multirow{2}{*}{Stage 3} & Latency (ms) & 6.6 & 1.6 \\
    & FLOPs (M) & 157.3 & 11.4 \\
    \hline
    \multirow{2}{*}{Stage 4} & Latency (ms) & 6.5 & 1.1 \\
    & FLOPs (M) & 157.3 & 2.8 \\
    \bottomrule
    \end{tabular}
    \label{table:channel_preserve}
    \vspace{-4mm}
\end{table}

In modern backbone architecture, The number of channels is continuously increasing from the bottom to the top layer \cite{szegedy2017inception, resnet, tan2019efficientnet}.
In ResNet, for example, the channel dimension is doubled when its spatial dimension is decreased (stride of 2).
This designing trend is based on that high-level features are highly related to the specific classes \cite{zeiler2014visualizing}.
When the number of object \JH{classes} increases, high-level feature dimension has to be enlarged accordingly.
However, there is only one object \JH{class} in the face detection task, so we suppose that the channel dimension may be reduced compared with the object detection task.

To accelerate face detection speed, EResNet backbone is designed by reducing the number of channels.
From the assumption, we propose the channel dimension preserving strategy, which means that we do not double the channel dimension for every stage.
Table~\ref{table:channel_preserve} shows the number of FLOPs and latency when our channel-preserving strategy is applied, and it shows that our method significantly reduced the latency and FLOPs of each stage.
\yj{We note that the number of FLOPs is also proportional to input and output channel dimension, and hence the amount of the reduction is huge.}
The latency reduction amount is \yj{not as large as} FLOPs reduction, but it is still remarkable.
\cam{Based on the intuition (Figure \ref{fig:conv_comparison}) that standard convolution is faster than depthwise separable convolution under the channel dimension less than 16,} we set the channel dimension of our EResNet\cam{-1x} architecture to 16.

The network capacity also decreases due to the channel dimension reduction, so we adjust the stage configuration to compensate for the performance degeneration.
\figurename~\ref{fig:eresfd_arch} shows the detailed stage configurations.
We insert one more block for stages 2$\sim$4 to improve small face detection performance and add two \JH{extra} stages (stages 5 and 6) for the large face.
The number of residual blocks increases from $8$ to $14$, so additional residual blocks can increase the inference time.
However, the channel-preserving strategy significantly reduces FLOPs, and the computational cost of each block is also much smaller than the original residual block.
For this reason, EResNet architecture is still much faster than the vanilla ResNet architecture \JH{as shown in Figure \ref{fig:latency}.}

\subsection{Feature Enhance Modules}
\subsubsection{SepFPN}

To improve the detection performance for small objects, feature pyramid network (FPN)~\cite{fpn} is widely adopted~\cite{faceboxes_improved, pyramidbox, deng2019retinaface}.
\cam{\yj{FPN propagates the context of high-level features from deep layers into low-level features from shallow layers (top-down), enriching the low-level features to better detect the small objects.}
However, previous work~\cite{pyramidbox} claimed that aggregating high-level features onto low-level features can hurt detection performance for small faces. This is because the large receptive field of the high-level features might convey irrelevant global contexts to the low-level features, obscuring their local contexts and thereby impeding their detection ability of small local faces.} 

\begin{figure}[t]
    \centering
    \includegraphics[width=0.5\textwidth]{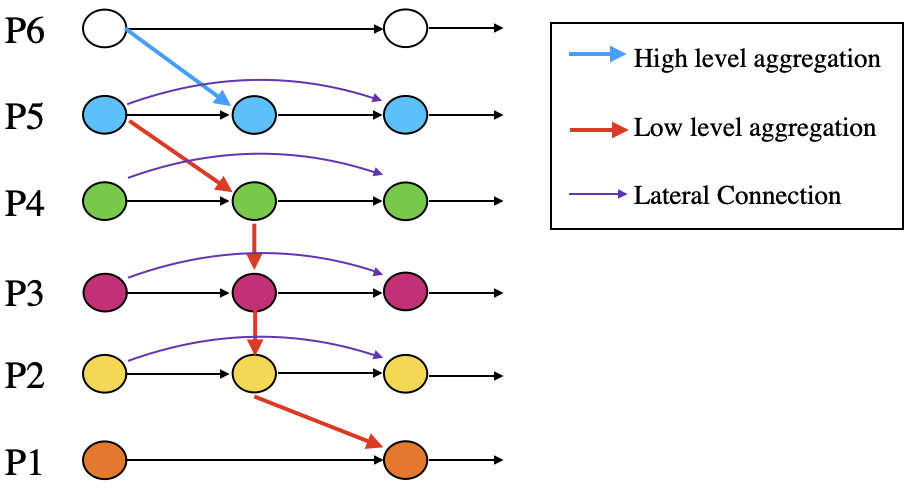}
    \caption{Architecture of SepFPN. \cam{P1$\sim$P6 denotes the intermediate features from low to high-level layers.} The high-level features and low-level features are aggregated separately.}
    \label{fig:sepfpn}
\end{figure}

\cam{To resolve this, we propose a new FPN module, separated feature pyramid network (SepFPN, Figure~\ref{fig:sepfpn}).}
\cam{From the above-mentioned observation \cite{pyramidbox}, we assume that a significant disparity of receptive field \yj{among the aggregation features} can lead to performance degradation. To address this, we separately organize the aggregation features in a hierarchical manner. Specifically, we ensure that high-level features are aggregated solely with other high-level features, while low-level features are exclusively combined with other low-level ones. Each of these two separated top-bottom paths shares similar contexts with similar sizes of receptive field within its aggregation group, consistently enhancing the detection ability across all the scales of faces.}

 \cam{For the aggregation details, we follow BiFPN~\cite{tan2020efficientdet} to aggregate features in a learnable manner with a simple element-wise weighted summation, where the latency overhead is negligible. We also introduce a lateral connection to avoid dilution of each original feature, as in BiFPN. Meanwhile, although BiFPN and several heavyweight object detectors~\cite{tan2020efficientdet, liu2018path} proposed to append an additional bottom-up aggregation path ($i.e.,$ from low to high-level features), we do not employ this scheme due to its large latency overhead.}




\subsubsection{CCPM}

\begin{figure}[t]
    \centering
    \includegraphics[width=0.4\textwidth]{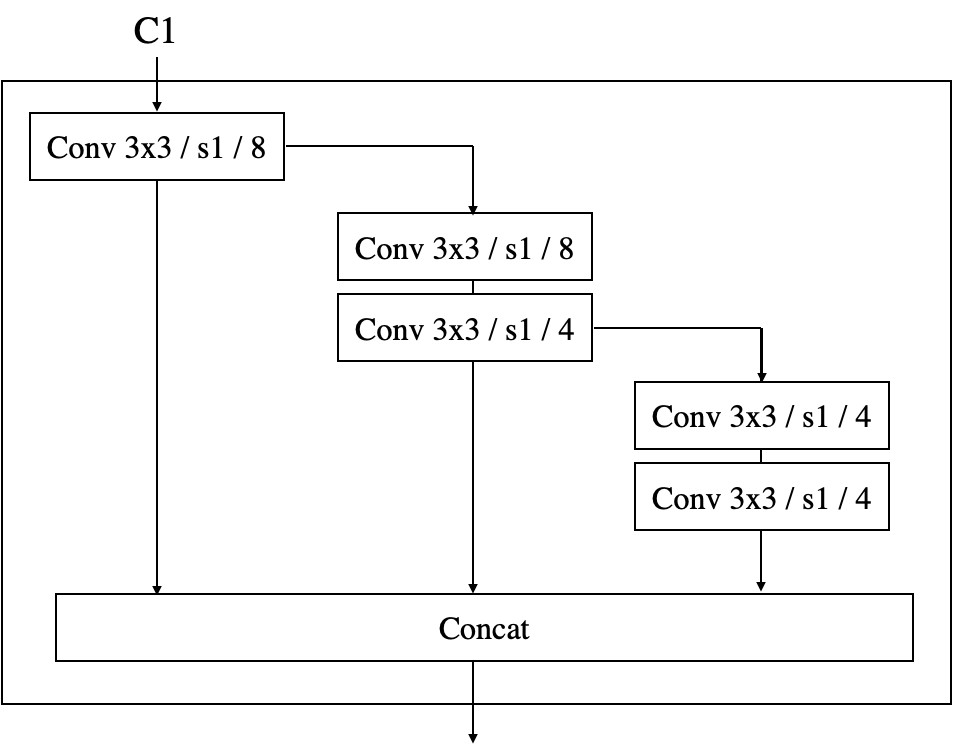}
    \caption{Architecture of CCPM in case of EResFD-1x. We only visualize CCPM for the feature map C1 in Figure \ref{fig:eresfd_arch} for simplicity.}
    \label{fig:ccpm}
\end{figure}

\cam{\yj{To \cam{further} supplement the feature information,} we also \cam{propose} cascade context prediction module (CCPM, Figure~\ref{fig:ccpm}) \cam{with a latency-aware module design}.
While the context prediction module~\cite{ssh, li2019pyramidbox++, refineface} was originally proposed to enlarge the receptive field, \yj{our cascade design of CCPM aims for the same objective but promotes faster latency.}
\JH{Specifically, our cascade structure can effectively enrich the large size of receptive field by reusing the previously convolved features, while ensuring faster speed than the previous heavyweight enhance modules using a large number of convolution layers~\cite{ssh}, densely-connected convolution layers~\cite{li2019pyramidbox++} and convolution layers with large asymmetric kernel~\cite{refineface}.}}
%
\cam{Owing to these advantages, CCPM helps to construct a highly efficient face detector that achieves high detection performance with low latency.}


\section{Experiment}
 \JH{In this section, we evaluate our proposed EResFD by analyzing the effectiveness of each component of EResFD and by comparing with the state-of-the-art (SOTA) face detectors. For quantitatively measuring the accuracy of detection, we used WIDER FACE~\cite{widerface} dataset. \JH{For training on WIDER FACE, color distortion, zoom-in and out augmentation, max-out background label, and multi-task loss are used, following S3FD~\cite{s3fd}. 
 For evaluation, we employed flip and multi-scale testing~\cite{s3fd}, where all these predictions are merged by Box voting~\cite{gidaris2015object} with intersection-over-union (IoU) threshold at 0.3. In the case of using RetinaFace framework~\cite{deng2019retinaface}\footnote{We obtained source code from \url{https://github.com/biubug6/Pytorch_Retinaface}}
}, we used single-scale testing where the original image size is maintained. %
 For measuring latency, we used Intel Xeon CPU (E5-2660v3@2.60 GHz) with VGA input resolution (480${\times}$640).}
\subsection{Component Study}
\paragraph{Backbone Network.}

\begin{figure}[t]
    \centering
    \begin{subfigure}[t]{0.48\linewidth}
        \includegraphics[width=1\linewidth]{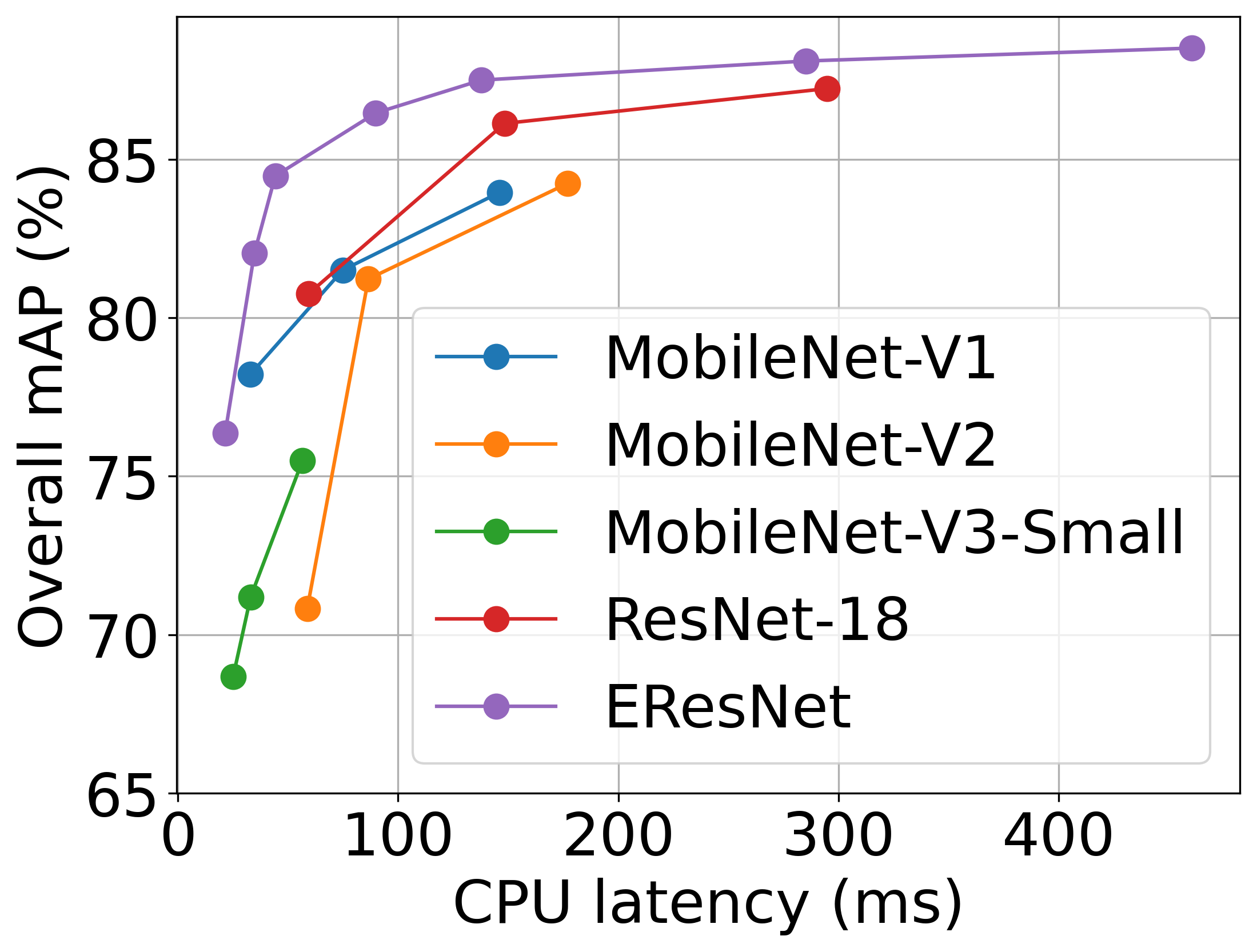}
        \caption{latency vs accuracy}
        \label{fig:latency_vs_accuracy}
    \end{subfigure}
    \begin{subfigure}[t]{0.48\linewidth}
        \includegraphics[width=1\linewidth]{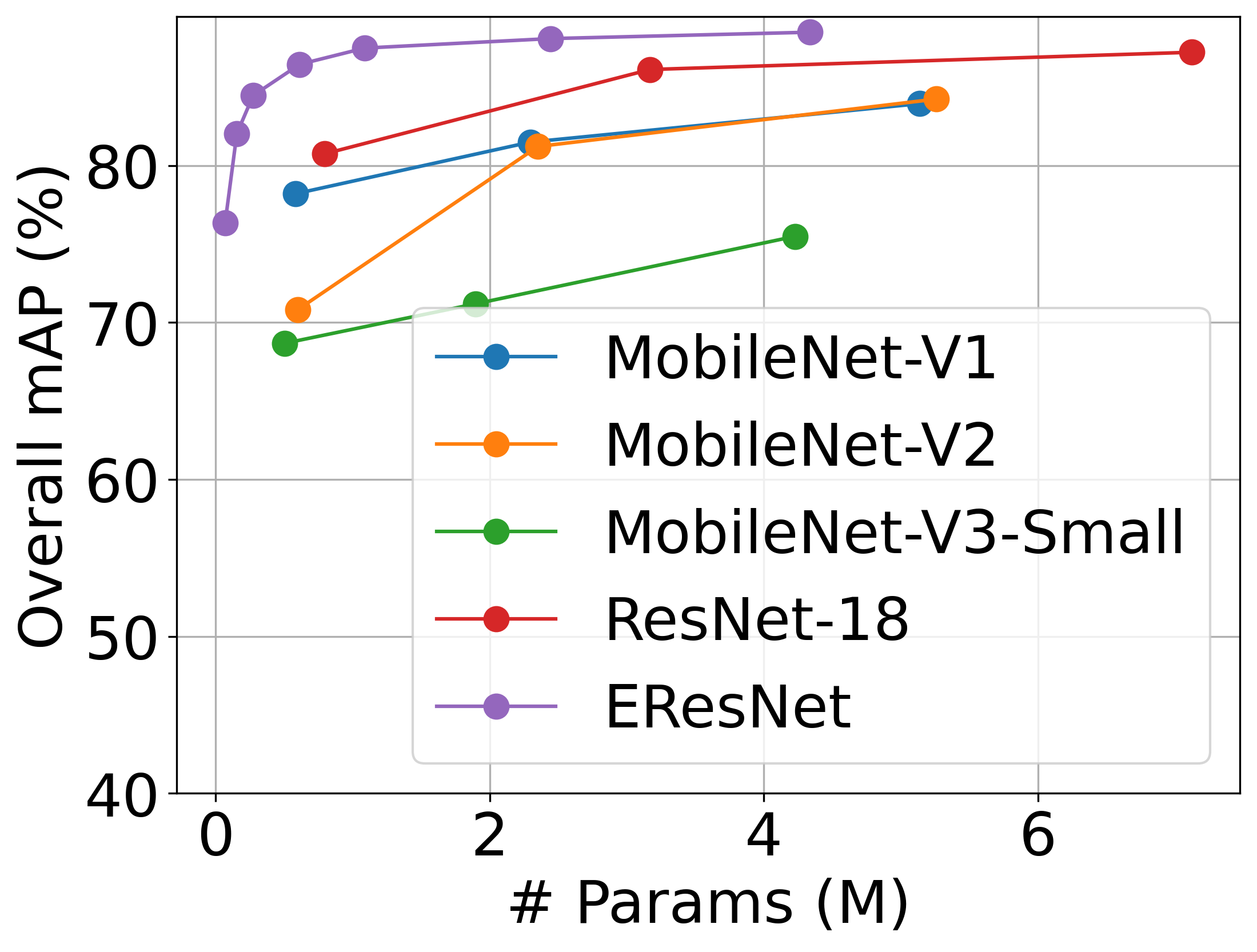}
        \centering
    \caption{parameter vs accuracy}
        \label{fig:params_vs_accuracy}
    \end{subfigure}
    \caption{Performance comparison of various backbone networks in terms of CPU latency and number of parameters. We applied width multipliers 1x, 1.5x, 2x, 3x, 4x, 6x, 8x for EResNet. For the other backbones, we applied width multipliers 0.25x, 0.5x, 0.75x.}
    \label{fig:backbone_comparison}
\end{figure}

\begin{table}[t]
    \small
    \centering
    \caption{Performance comparison of various FPN modules on WIDER FACE.
    }
    \smallskip
    \noindent
    \resizebox{1.05\linewidth}{!}{
    \begin{tabular}{c|c|cccc}
    \toprule
    \multicolumn{1}{c|}{\multirow{2}{*}{Model}} & \multirow{2}{*}{Latency (ms)} & \multicolumn{4}{c}{mAP (\%)}                                                                   \\ \cline{3-6} 
    \multicolumn{1}{c|}{}                  &                               & \multicolumn{1}{c}{Easy}      & \multicolumn{1}{c}{Medium} & \multicolumn{1}{c|}{Hard}  & Overall \\ \hline
    EResNet                                 & 20.9                          & \multicolumn{1}{c}{85.09} & \multicolumn{1}{c}{82.78}  & \multicolumn{1}{c|}{61.20}  & 76.36   \\ \hline
    + FPN~\cite{fpn}                                   & 24.0                          & \multicolumn{1}{c}{85.30}  & \multicolumn{1}{c}{84.25}  & \multicolumn{1}{c|}{75.45} & 81.67   \\ \hline
    + LFPN~\cite{pyramidbox}                                  & 23.8                          & \multicolumn{1}{c}{85.24} & \multicolumn{1}{c}{84.21}  & \multicolumn{1}{c|}{76.58} & 82.01   \\ \hline
    + PANet~\cite{liu2018path}                             & 35.5                          & \multicolumn{1}{c}{87.96} & \multicolumn{1}{c}{86.82}  & \multicolumn{1}{c|}{77.95} & 84.24   \\ \hline
    + BiFPN~\cite{tan2020efficientdet}                                 & 35.6                          & \multicolumn{1}{c}{87.16} & \multicolumn{1}{c}{85.95}  & \multicolumn{1}{c|}{77.56} & 83.56   \\ \hline
    + SepFPN (Ours)                               & \textbf{27.0}                          & \multicolumn{1}{c}{87.68} & \multicolumn{1}{c}{86.30}   & \multicolumn{1}{c|}{77.68} & 83.89   \\ 
    \bottomrule
    \end{tabular}
    }
    \label{table:fpn}
\end{table}


\figurename~\ref{fig:backbone_comparison} shows the comparison result with widely used lightweight backbone; ResNet18, MobileNetV1, V2, and V3. 
\yj{The experimental results show that EResNet achieves superior inference latency given the similar mAP condition} and also has higher mAP given the similar latency condition, as shown in \figurename~\ref{fig:latency_vs_accuracy}.
Our proposed stem layer and channel dimension preserving strategy \yj{are shown to be very helpful for latency reduction, while maintaining the powerful face detection performance. }
In addition, EResNet also outperforms other comparison methods in terms of the number of parameters.
In \figurename~\ref{fig:params_vs_accuracy}, EResNet shows the highest mAP with a much smaller number of parameters.
\JH{To further prove the general effectiveness of EResNet backbone, we additionally compared it with various backbone architectures on RetinaFace framework in Figure \ref{fig:latency}. 
For all the backbones, we only employed 3 detection heads from P2 $\sim$ P4 in Figure \ref{fig:eresfd_arch}, following \cite{deng2019retinaface}. 
The results further corroborate that our EResNet architecture has the best latency-accuracy trade-off among the various backbones.}
From those experiments, we found that the proposed methods effectively reduce both latency and parameters without causing mAP degeneration.

\paragraph{SepFPN.}
\begin{table}[t]
    \small
    \centering
    \caption{Ablation study of SepFPN with various separation position. In case of separation position is 5 (Figure \ref{fig:sepfpn}), high level features are only aggregated from P6 to P5 and the rest low-level features are aggregated from P5 to P1.
    }
    \smallskip
    \noindent
    \resizebox{1.05\linewidth}{!}{
        \begin{tabular}{c|c|cccc}
        \toprule
        \multirow{2}{*}{Separation Position} & \multirow{2}{*}{Latency (ms)} & \multicolumn{4}{c}{mAP (\%)}                                                                   \\ \cline{3-6} 
                                             &                               & \multicolumn{1}{c}{Easy}  & \multicolumn{1}{c}{Medium} & \multicolumn{1}{c|}{Hard}  & Overall \\ \hline
        P3                                   & 26.9                          & \multicolumn{1}{c}{85.95} & \multicolumn{1}{c}{84.39}  & \multicolumn{1}{c|}{75.31} & 81.88   \\ \hline
        P4                                   & 26.7                          & \multicolumn{1}{c}{87.16} & \multicolumn{1}{c}{85.70}  & \multicolumn{1}{c|}{77.12} & 83.33   \\ \hline
        P5                                   & 27.0                          & \multicolumn{1}{c}{87.68} & \multicolumn{1}{c}{86.30}  & \multicolumn{1}{c|}{77.68} & 83.89   \\ 
        \bottomrule
        \end{tabular}
    }
    \label{table:sepfpn_ablation}
\end{table}
\cam{We measured the detection performance and latency of several other FPN modules on the EResNet backbone, in Table~\ref{table:fpn}.
Compared to the lightweight FPN modules such as FPN and LFPN, our SepFPN achieves much more detection accuracy gain with a small increase of latency. Meanwhile, compared to the heavyweight FPN modules ($i.e.,$ PANet, BiFPN) with bottom-up aggregation path, our SepFPN achieves comparable or even higher detection performance, while exhibiting $24\%$ shortened latency.}
This experiment shows that the bottom-up path \yj{would not be an essential block} for efficient face detection. 
We further empirically studied on the separation position of SepFPN, and Table~\ref{table:sepfpn_ablation} shows the result.
The latency is not significantly affected by the separation position, but the accuracy is very sensitive according to the separation position.
We observed that P5 achieves the best mAP for all different kinds of face sizes, and hence we applied P5 for all other experiments.

\paragraph{CCPM.}
\begin{table}[t]
    \small
    \centering
    \caption{Performance comparison of various feature enhance modules before detection head on WIDER FACE. For fair comparisons, we fix baseline backbone network as EResNet-1x equipped with LFPN~\cite{pyramidbox}.}
    \smallskip
    \noindent
    \resizebox{1.05\linewidth}{!}{
    \begin{tabular}{c|c|cccc}
    \toprule
    \multicolumn{1}{c|}{\multirow{2}{*}{Model}} & \multirow{2}{*}{Latency (ms)} & \multicolumn{4}{c}{mAP (\%)}                                                                   \\ \cline{3-6} 
    \multicolumn{1}{c|}{}                  &                               & \multicolumn{1}{c}{Easy}      & \multicolumn{1}{c}{Medium} & \multicolumn{1}{c|}{Hard}  & Overall \\ \hline
    Baseline                                 & 23.8          & \multicolumn{1}{c}{85.24} & \multicolumn{1}{c}{84.21}  & \multicolumn{1}{c|}{76.58}  & 82.01   \\ \hline
    + SSH~\cite{ssh}                                 & 35.5                          & \multicolumn{1}{c}{87.49} & \multicolumn{1}{c}{86.34}  & \multicolumn{1}{c|}{79.28}  & 84.37   \\ \hline
    + CPM~\cite{pyramidbox}                                 & 42.4                          & \multicolumn{1}{c}{87.47} & \multicolumn{1}{c}{86.74}  & \multicolumn{1}{c|}{80.00}  &  84.74  \\ \hline
    + FEM~\cite{dsfd}                                 & 41.5                          & \multicolumn{1}{c}{86.90} & \multicolumn{1}{c}{86.15}  & \multicolumn{1}{c|}{79.22}  & 84.09   \\ \hline
    + DCM~\cite{li2019pyramidbox++}                                 & 48.1                          & \multicolumn{1}{c}{87.48} & \multicolumn{1}{c}{86.51}  & \multicolumn{1}{c|}{79.85}  &  84.61 \\ \hline
  + CCPM (ours)                                 & \textbf{33.8}                          & \multicolumn{1}{c}{87.25} & \multicolumn{1}{c}{86.38}  & \multicolumn{1}{c|}{79.90}  &  84.51 \\ 
    \bottomrule
    \end{tabular}
    }
    \label{table:cpm}
\end{table}

Table \ref{table:cpm} shows a performance comparison result of various feature enhance modules. 
The feature enhance module makes large performance gain, but several previous works~\cite{pyramidbox, dsfd, li2019pyramidbox++} show similar detection performance.
Our CCPM module mainly focuses on latency reduction, and experimental result shows that CCPM achieves the fastest latency, satisfying the purpose.
In addition, CCPM also achieves higher overall mAP than SSH, which is the fastest among all the previous methods mentioned in the Table.


\subsection{Comparison with SOTA Detectors}


\begin{table*}[t]
    \small
    \centering
    \caption{Comparison with previous works on WIDER FACE validation set. All models are evaluated with multi-scale testing, following \cite{s3fd, deng2019retinaface}. For measuring FLOPs and Latency, VGA resolution (480${\times}$640) is used. For MTCNN, we used input sizes designated by \cite{mtcnn}.  
    }
    \vspace{-3mm}
    \smallskip
    \noindent
    \resizebox{0.95\linewidth}{!}{
    \begin{tabular}{c|cc|ccc|ccc|c}
    \toprule
    \multirow{2}{*}{Method} & \multirow{2}{*}{Backbone} & \multirow{2}{*}{Feature Enhance Module} & \multirow{2}{*}{\# Params} & \multirow{2}{*}{\# FLOPs} & \multirow{2}{*}{Latency} & \multicolumn{4}{c}{mAP (\%)}                                                                   \\ \cline{7-10} 
    & & & & & & \multicolumn{1}{c}{Easy} & \multicolumn{1}{c}{Medium} & \multicolumn{1}{c|}{Hard}  & Overall \\ \hline 
    MTCNN~\cite{mtcnn} & P-,R-,O-Net~\cite{mtcnn} & - & 0.12M & 14M & 4.0ms & 85.10 & 82.00 & 60.70 & 75.93 \\
    FaceBoxes~\cite{faceboxes_improved} & FaceBoxes~\cite{faceboxes_improved} & FPN + DCH & 0.66M & 156M & 35.7ms & 88.50 & 86.20 & 77.30 & 84.00 \\
    RetinaFace~\cite{deng2019retinaface} & MobileNetV1-0.25x~\cite{howard2017mobilenets} & FPN + SSH & 0.42M & 754M & 58.5ms & 88.67 & 87.09 & \textbf{80.99} & 85.58 \\ \hline 
    EResFD & \textbf{EResNet-1x} & - & 0.07M & 228M & 20.9ms & 85.09 & 82.78 & 61.20 & 76.36\\
    EResFD & \textbf{EResNet-1x} & \textbf{SepFPN} & 0.08M & 250M & 27.0ms & 87.68 & 86.30 & 77.68 & 83.89 \\
    EResFD & \textbf{EResNet-1x} & \textbf{SepFPN} + \textbf{CCPM} & 0.09M & 298M & 37.7ms & \textbf{89.02} & \textbf{87.96} & 80.41 & \textbf{85.80} \\ 
    \bottomrule
    \end{tabular}
    }
    \label{table:sota_comp}
\end{table*}

 We compare our proposed method with the SOTA real-time CPU detectors on WIDER FACE validation dataset.
Table~\ref{table:sota_comp} (upper part) shows the comparison result.
\textbf{EResNet-1x} indicates EResNet backbone architecture shown in \figurename~\ref{fig:eresfd_arch}, with width multiplier 1.
In the case of RetinaFace \cite{deng2019retinaface} backbone, width multiplier 0.25 is applied. 
MTCNN shows the smallest FLOPs and Latency, but it has large mAP degradation for medium and hard case.
EResFD has the smallest number of parameters and also achieves the highest overall mAP.
The latency of EResFD is similar to that of FaceBoxes \cite{zhang2019faceboxes}, but its detection performance is much higher.
Moreover, the proposed method achieves similar or slightly higher mAP compared with RetinaFace \cite{deng2019retinaface}, but its latency is about $64$\% of that of RetinaFace.

Table~\ref{table:sota_comp} (bottom part) also shows the ablation study for the proposed modules; SepFPN and CCPM.
SepFPN improves the overall mAP by about $7.5$\%, but its latency only increases by $6$ \textit{ms}.
Moreover, we also achieve $1.9$\% overall mAP improvement when CCPM is further applied.
We observed that proposed modules can make a large performance improvement even when jointly applied.
\begin{figure}
\centering
    \includegraphics[width=\columnwidth]{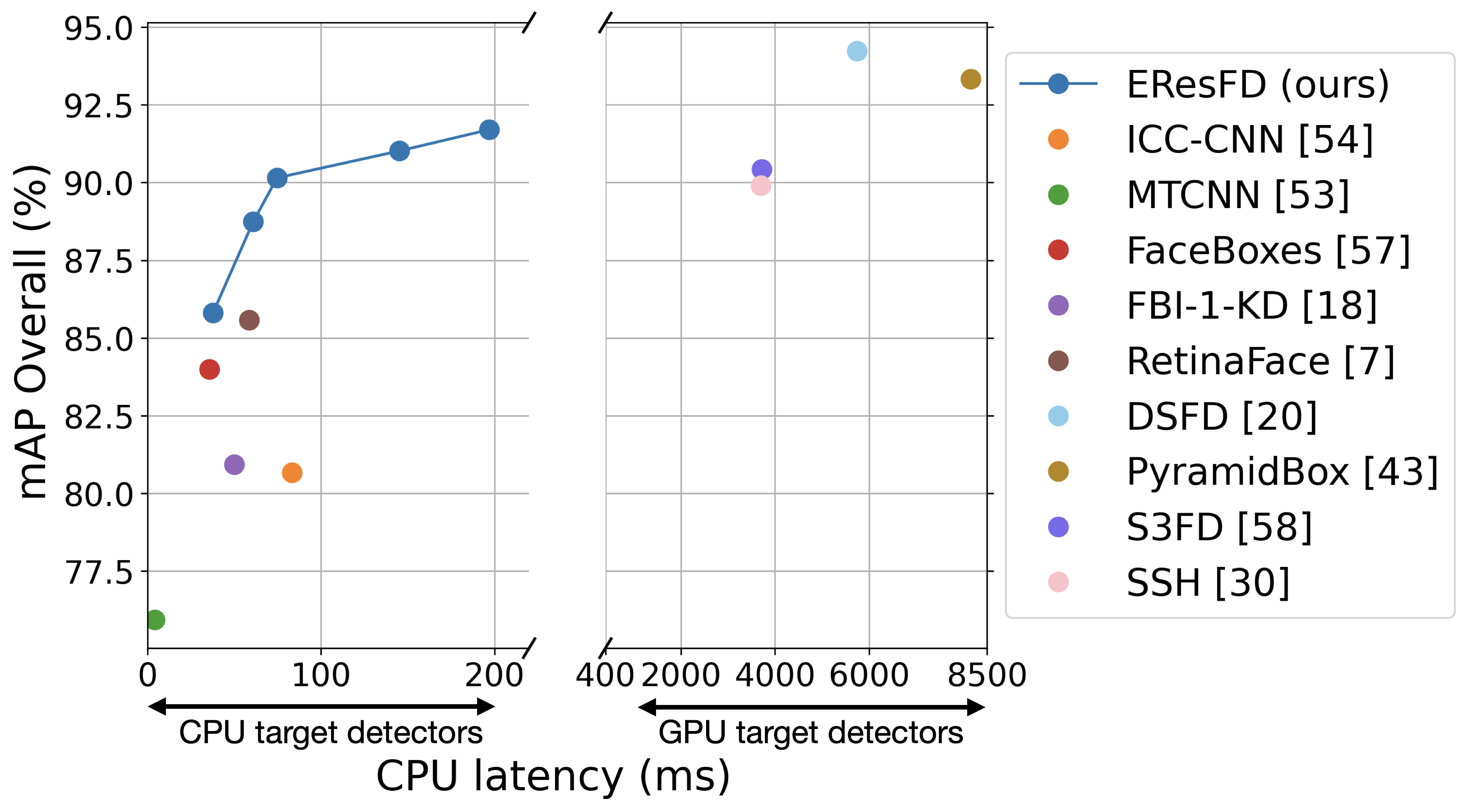}
	\vspace{-3mm}
	\caption{Performance comparison of EResFD with other SOTA CPU target detectors~\cite{zhang2017detecting, mtcnn, faceboxes_improved, jin2019learning, deng2019retinaface} and GPU target detectors~\cite{dsfd, pyramidbox, s3fd, ssh}. For RetinaFace~\cite{deng2019retinaface},
	MobileNetV1-0.25x backbone was used.}
	\vspace{-6mm}
	\label{fig:comparison_with_cpu_gpu_detectors}
\end{figure}

In addition, we also compare the latency and detection performance with other CPU and GPU target face detectors in \figurename~\ref{fig:comparison_with_cpu_gpu_detectors}.
As we already mentioned above, our method achieves the highest mAP among all the CPU target face detectors.
In addition, we found that it also shows comparable detection performance compared with GPU target detectors.
EResFD shows similar detection accuracy with S3FD and SSH, but it is about $19$x faster.

\begin{figure}[t]
    \centering
    \begin{subfigure}[t]{0.45\linewidth}
        \centering
        \includegraphics[width=1\linewidth]{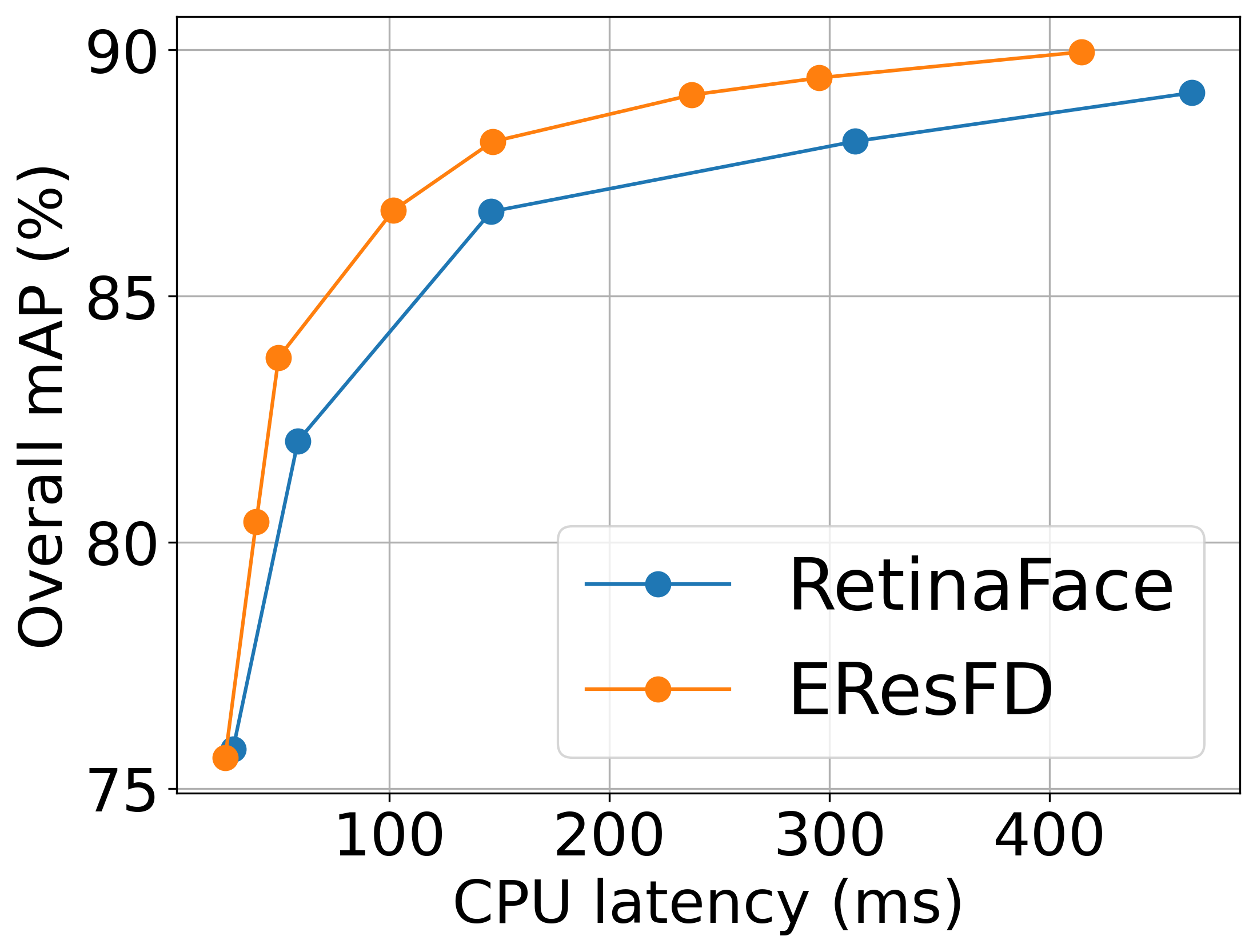}
        \subcaption{Detection}
        \vspace{3mm}
    \end{subfigure}
    \begin{subfigure}[t]{0.45\linewidth}
        \includegraphics[width=1\linewidth]{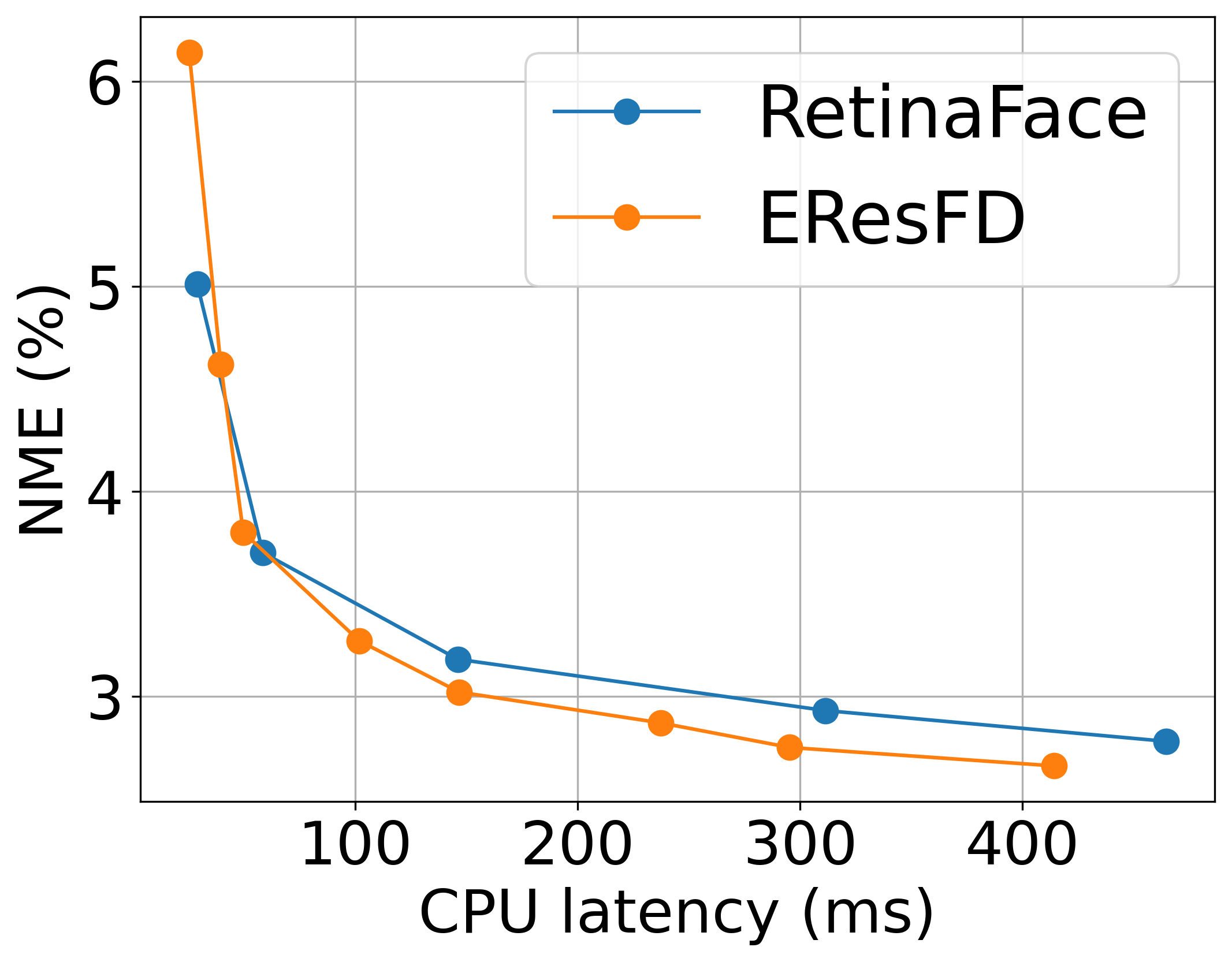}
        \centering
    \subcaption{Landmark detection}
    \end{subfigure}
    \vspace{-3mm}
    \caption{Performance of face detection on WIDER FACE and landmark detection on AFLW~\cite{koestinger2011annotated} dataset. Based on RetinaFace framework where face bounding boxes with facial landmarks can be jointly detected, we only replaced the backbone network from MobileNetV1 to EResNet, while FPN and SSH are replaced by our proposed SepFPN and CCPM, respectively for EResFD.}
    \vspace{-6mm}
    \label{fig:retinaface_comparison}
\end{figure}

Furthermore, we compare the proposed method with RetinaFace, \yj{one standard lightweight face detector in this field.}
Since RetinaFace detects face and facial landmarks at the same time, we covered landmark detection as well.
We measured the face detection performance on WIDER FACE and landmark detection performance on AFLW.
\figurename~\ref{fig:retinaface_comparison} shows comparison on face and landmark detection.
Our EResFD achieves higher mAP for face detection and lower NME for landmark detection, while reducing latency.
\section{Conclusion}
This paper rediscovers the efficiency of standard convolution-based architecture for lightweight face detection. The extensive experimental results showed that the standard convolutional block achieves superior performance compared to depthwise separable convolution, contrary to the common trend in this field. Based on the observation, we propose an efficient architecture EResNet, which includes a modified stem layer and channel dimension preserving strategy. Also, we propose SepFPN and CCPM for the feature enhancement, which boosts the detection performance without sacrificing latency and parameter size. Summing up the observations and architecture suggestions for face detection, we establish a new state-of-the-art real-time CPU face detector, EResFD, achieving the SOTA face detection performance among the lightweight detectors.

{\small
\bibliographystyle{ieee_fullname}
\bibliography{egbib}

\begin{thebibliography}{10}\itemsep=-1pt

\bibitem{bazarevsky2019blazeface}
Valentin Bazarevsky, Yury Kartynnik, Andrey Vakunov, Karthik Raveendran, and
  Matthias Grundmann.
\newblock Blazeface: Sub-millisecond neural face detection on mobile gpus.
\newblock {\em arXiv preprint arXiv:1907.05047}, 2019.

\bibitem{bello2021revisiting}
Irwan Bello, William Fedus, Xianzhi Du, Ekin~Dogus Cubuk, Aravind Srinivas,
  Tsung-Yi Lin, Jonathon Shlens, and Barret Zoph.
\newblock Revisiting resnets: Improved training and scaling strategies.
\newblock {\em Advances in Neural Information Processing Systems}, 34, 2021.

\bibitem{cai2018cascade(cascadercnn)}
Zhaowei Cai and Nuno Vasconcelos.
\newblock Cascade r-cnn: Delving into high quality object detection.
\newblock In {\em Proceedings of the IEEE conference on computer vision and
  pattern recognition}, pages 6154--6162, 2018.

\bibitem{cao2021emface}
Leilei Cao, Yao Xiao, and Lin Xu.
\newblock Emface: Detecting hard faces by exploring receptive field pyraminds.
\newblock {\em arXiv preprint arXiv:2105.10104}, 2021.

\bibitem{chen2018encoder}
Liang-Chieh Chen, Yukun Zhu, George Papandreou, Florian Schroff, and Hartwig
  Adam.
\newblock Encoder-decoder with atrous separable convolution for semantic image
  segmentation.
\newblock In {\em Proceedings of the European conference on computer vision
  (ECCV)}, pages 801--818, 2018.

\bibitem{deng2009imagenet}
Jia Deng, Wei Dong, Richard Socher, Li-Jia Li, Kai Li, and Li Fei-Fei.
\newblock Imagenet: A large-scale hierarchical image database.
\newblock In {\em 2009 IEEE conference on computer vision and pattern
  recognition}, pages 248--255. Ieee, 2009.

\bibitem{deng2019retinaface}
Jiankang Deng, Jia Guo, Yuxiang Zhou, Jinke Yu, Irene Kotsia, and Stefanos
  Zafeiriou.
\newblock Retinaface: Single-stage dense face localisation in the wild.
\newblock {\em arXiv preprint arXiv:1905.00641}, 2019.

\bibitem{gidaris2015object}
Spyros Gidaris and Nikos Komodakis.
\newblock Object detection via a multi-region and semantic segmentation-aware
  cnn model.
\newblock In {\em Proceedings of the IEEE international conference on computer
  vision}, pages 1134--1142, 2015.

\bibitem{guo2021sample}
Jia Guo, Jiankang Deng, Alexandros Lattas, and Stefanos Zafeiriou.
\newblock Sample and computation redistribution for efficient face detection.
\newblock In {\em International Conference on Learning Representations}, 2022.

\bibitem{han2022learning}
Dongyoon Han, YoungJoon Yoo, Beomyoung Kim, and Byeongho Heo.
\newblock Learning features with parameter-free layers.
\newblock {\em arXiv preprint arXiv:2202.02777}, 2022.

\bibitem{han2020ghostnet}
Kai Han, Yunhe Wang, Qi Tian, Jianyuan Guo, Chunjing Xu, and Chang Xu.
\newblock Ghostnet: More features from cheap operations.
\newblock In {\em Proceedings of the IEEE/CVF Conference on Computer Vision and
  Pattern Recognition}, pages 1580--1589, 2020.

\bibitem{he2017mask(mrcnn)}
Kaiming He, Georgia Gkioxari, Piotr Doll{\'a}r, and Ross Girshick.
\newblock Mask r-cnn.
\newblock In {\em Proceedings of the IEEE international conference on computer
  vision}, pages 2961--2969, 2017.

\bibitem{resnet}
Kaiming He, Xiangyu Zhang, Shaoqing Ren, and Jian Sun.
\newblock Deep residual learning for image recognition.
\newblock In {\em Proceedings of the IEEE conference on computer vision and
  pattern recognition}, pages 770--778, 2016.

\bibitem{hoang2020deface}
Toan~Minh Hoang, Gi~Pyo Nam, Junghyun Cho, and Ig-Jae Kim.
\newblock Deface: Deep efficient face network for small scale variations.
\newblock {\em IEEE Access}, 8:142423--142433, 2020.

\bibitem{howard2019searching}
Andrew Howard, Mark Sandler, Grace Chu, Liang-Chieh Chen, Bo Chen, Mingxing
  Tan, Weijun Wang, Yukun Zhu, Ruoming Pang, Vijay Vasudevan, et~al.
\newblock Searching for mobilenetv3.
\newblock In {\em Proceedings of the IEEE/CVF International Conference on
  Computer Vision}, pages 1314--1324, 2019.

\bibitem{howard2017mobilenets(mobilenet)}
Andrew~G Howard, Menglong Zhu, Bo Chen, Dmitry Kalenichenko, Weijun Wang,
  Tobias Weyand, Marco Andreetto, and Hartwig Adam.
\newblock Mobilenets: Efficient convolutional neural networks for mobile vision
  applications.
\newblock {\em arXiv preprint arXiv:1704.04861}, 2017.

\bibitem{howard2017mobilenets}
Andrew~G Howard, Menglong Zhu, Bo Chen, Dmitry Kalenichenko, Weijun Wang,
  Tobias Weyand, Marco Andreetto, and Hartwig Adam.
\newblock Mobilenets: Efficient convolutional neural networks for mobile vision
  applications.
\newblock {\em arXiv preprint arXiv:1704.04861}, 2017.

\bibitem{jin2019learning}
Haibo Jin, Shifeng Zhang, Xiangyu Zhu, Yinhang Tang, Zhen Lei, and Stan~Z Li.
\newblock Learning lightweight face detector with knowledge distillation.
\newblock In {\em 2019 International Conference on Biometrics (ICB)}, pages
  1--7. IEEE, 2019.

\bibitem{koestinger2011annotated}
Martin Koestinger, Paul Wohlhart, Peter~M Roth, and Horst Bischof.
\newblock Annotated facial landmarks in the wild: A large-scale, real-world
  database for facial landmark localization.
\newblock In {\em 2011 IEEE international conference on computer vision
  workshops (ICCV workshops)}, pages 2144--2151. IEEE, 2011.

\bibitem{dsfd}
Jian Li, Yabiao Wang, Changan Wang, Ying Tai, Jianjun Qian, Jian Yang, Chengjie
  Wang, Jilin Li, and Feiyue Huang.
\newblock Dsfd: dual shot face detector.
\newblock In {\em Proceedings of the IEEE Conference on Computer Vision and
  Pattern Recognition}, pages 5060--5069, 2019.

\bibitem{li2018dsfd}
Jian Li, Yabiao Wang, Changan Wang, Ying Tai, Jianjun Qian, Jian Yang, Chengjie
  Wang, Jilin Li, and Feiyue Huang.
\newblock Dsfd: Dual shot face detector.
\newblock In {\em Proceedings of the IEEE Conference on Computer Vision and
  Pattern Recognition}, 2019.

\bibitem{li2019pyramidbox++}
Zhihang Li, Xu Tang, Junyu Han, Jingtuo Liu, and Ran He.
\newblock Pyramidbox++: high performance detector for finding tiny face.
\newblock {\em arXiv preprint arXiv:1904.00386}, 2019.

\bibitem{lin2021mcunetv2}
Ji Lin, Wei-Ming Chen, Han Cai, Chuang Gan, and Song Han.
\newblock Mcunetv2: Memory-efficient patch-based inference for tiny deep
  learning.
\newblock {\em arXiv preprint arXiv:2110.15352}, 2021.

\bibitem{fpn}
Tsung-Yi Lin, Piotr Doll{\'a}r, Ross Girshick, Kaiming He, Bharath Hariharan,
  and Serge Belongie.
\newblock Feature pyramid networks for object detection.
\newblock In {\em Proceedings of the IEEE conference on computer vision and
  pattern recognition}, pages 2117--2125, 2017.

\bibitem{lin2017focal(retinanet)}
Tsung-Yi Lin, Priya Goyal, Ross Girshick, Kaiming He, and Piotr Doll{\'a}r.
\newblock Focal loss for dense object detection.
\newblock In {\em Proceedings of the IEEE international conference on computer
  vision}, pages 2980--2988, 2017.

\bibitem{liu2018path}
Shu Liu, Lu Qi, Haifang Qin, Jianping Shi, and Jiaya Jia.
\newblock Path aggregation network for instance segmentation.
\newblock In {\em Proceedings of the IEEE conference on computer vision and
  pattern recognition}, pages 8759--8768, 2018.

\bibitem{ssd}
Wei Liu, Dragomir Anguelov, Dumitru Erhan, Christian Szegedy, Scott Reed,
  Cheng-Yang Fu, and Alexander~C Berg.
\newblock Ssd: Single shot multibox detector.
\newblock In {\em European conference on computer vision}, pages 21--37.
  Springer, 2016.

\bibitem{liu2020bfbox}
Yang Liu and Xu Tang.
\newblock Bfbox: Searching face-appropriate backbone and feature pyramid
  network for face detector.
\newblock In {\em Proceedings of the IEEE/CVF Conference on Computer Vision and
  Pattern Recognition}, pages 13568--13577, 2020.

\bibitem{liu2020hambox}
Yang Liu, Xu Tang, Junyu Han, Jingtuo Liu, Dinger Rui, and Xiang Wu.
\newblock Hambox: Delving into mining high-quality anchors on face detection.
\newblock In {\em 2020 IEEE/CVF Conference on Computer Vision and Pattern
  Recognition (CVPR)}, pages 13043--13051. IEEE, 2020.

\bibitem{ssh}
Mahyar Najibi, Pouya Samangouei, Rama Chellappa, and Larry~S Davis.
\newblock Ssh: Single stage headless face detector.
\newblock In {\em Proceedings of the IEEE International Conference on Computer
  Vision}, pages 4875--4884, 2017.

\bibitem{najibi2019fa}
Mahyar Najibi, Bharat Singh, and Larry~S Davis.
\newblock Fa-rpn: Floating region proposals for face detection.
\newblock In {\em Proceedings of the IEEE/CVF Conference on Computer Vision and
  Pattern Recognition}, pages 7723--7732, 2019.

\bibitem{qi2023fast}
Shuaihui Qi, Xiaofeng Song, Zhiyuan Li, and Tao Xie.
\newblock Fast and efficient face detector based on large kernel attention for
  cpu device.
\newblock {\em Journal of Real-Time Image Processing}, 20(4):1--11, 2023.

\bibitem{ramos2020swiftface}
Leonardo Ramos and Bernardo Morales.
\newblock Swiftface: Real-time face detection.
\newblock {\em arXiv preprint arXiv:2009.13743}, 2020.

\bibitem{redmon2016you(yolo)}
Joseph Redmon, Santosh Divvala, Ross Girshick, and Ali Farhadi.
\newblock You only look once: Unified, real-time object detection.
\newblock In {\em Proceedings of the IEEE conference on computer vision and
  pattern recognition}, pages 779--788, 2016.

\bibitem{ren2015faster}
Shaoqing Ren, Kaiming He, Ross Girshick, and Jian Sun.
\newblock Faster r-cnn: Towards real-time object detection with region proposal
  networks.
\newblock {\em Advances in neural information processing systems}, 28:91--99,
  2015.

\bibitem{sandler2018mobilenetv2}
Mark Sandler, Andrew Howard, Menglong Zhu, Andrey Zhmoginov, and Liang-Chieh
  Chen.
\newblock Mobilenetv2: Inverted residuals and linear bottlenecks.
\newblock In {\em Proceedings of the IEEE conference on computer vision and
  pattern recognition}, pages 4510--4520, 2018.

\bibitem{song2020kpnet}
Guanglu Song, Yu Liu, Yuhang Zang, Xiaogang Wang, Biao Leng, and Qingsheng
  Yuan.
\newblock Kpnet: Towards minimal face detector.
\newblock In {\em Proceedings of the AAAI Conference on Artificial
  Intelligence}, volume~34, pages 12015--12022, 2020.

\bibitem{szegedy2017inception}
Christian Szegedy, Sergey Ioffe, Vincent Vanhoucke, and Alexander~A Alemi.
\newblock Inception-v4, inception-resnet and the impact of residual connections
  on learning.
\newblock In {\em Thirty-first AAAI conference on artificial intelligence},
  2017.

\bibitem{tan2019efficientnet}
Mingxing Tan and Quoc Le.
\newblock Efficientnet: Rethinking model scaling for convolutional neural
  networks.
\newblock In {\em International Conference on Machine Learning}, pages
  6105--6114. PMLR, 2019.

\bibitem{tan2021efficientnetv2}
Mingxing Tan and Quoc~V Le.
\newblock Efficientnetv2: Smaller models and faster training.
\newblock {\em arXiv preprint arXiv:2104.00298}, 2021.

\bibitem{tan2020efficientdet(efficientdet)}
Mingxing Tan, Ruoming Pang, and Quoc~V Le.
\newblock Efficientdet: Scalable and efficient object detection.
\newblock In {\em Proceedings of the IEEE/CVF conference on computer vision and
  pattern recognition}, pages 10781--10790, 2020.

\bibitem{tan2020efficientdet}
Mingxing Tan, Ruoming Pang, and Quoc~V Le.
\newblock Efficientdet: Scalable and efficient object detection.
\newblock In {\em Proceedings of the IEEE/CVF conference on computer vision and
  pattern recognition}, pages 10781--10790, 2020.

\bibitem{pyramidbox}
Xu Tang, Daniel~K Du, Zeqiang He, and Jingtuo Liu.
\newblock Pyramidbox: A context-assisted single shot face detector.
\newblock In {\em Proceedings of the European Conference on Computer Vision
  (ECCV)}, pages 797--813, 2018.

\bibitem{vesdapunt2021crface}
Noranart Vesdapunt and Baoyuan Wang.
\newblock Crface: Confidence ranker for model-agnostic face detection
  refinement.
\newblock In {\em Proceedings of the IEEE/CVF Conference on Computer Vision and
  Pattern Recognition}, pages 1674--1684, 2021.

\bibitem{wu2018shift}
Bichen Wu, Alvin Wan, Xiangyu Yue, Peter Jin, Sicheng Zhao, Noah Golmant, Amir
  Gholaminejad, Joseph Gonzalez, and Kurt Keutzer.
\newblock Shift: A zero flop, zero parameter alternative to spatial
  convolutions.
\newblock In {\em Proceedings of the IEEE Conference on Computer Vision and
  Pattern Recognition}, pages 9127--9135, 2018.

\bibitem{wu2023yunet}
Wei Wu, Hanyang Peng, and Shiqi Yu.
\newblock Yunet: A tiny millisecond-level face detector.
\newblock {\em Machine Intelligence Research}, pages 1--10, 2023.

\bibitem{widerface}
Shuo Yang, Ping Luo, Chen-Change Loy, and Xiaoou Tang.
\newblock Wider face: A face detection benchmark.
\newblock In {\em Proceedings of the IEEE conference on computer vision and
  pattern recognition}, pages 5525--5533, 2016.

\bibitem{yoo2019extd}
YoungJoon Yoo, Dongyoon Han, and Sangdoo Yun.
\newblock Extd: Extremely tiny face detector via iterative filter reuse.
\newblock {\em arXiv preprint arXiv:1906.06579}, 2019.

\bibitem{zeiler2014visualizing}
Matthew~D Zeiler and Rob Fergus.
\newblock Visualizing and understanding convolutional networks.
\newblock In {\em European conference on computer vision}, pages 818--833.
  Springer, 2014.

\bibitem{zhang2020asfd}
Bin Zhang, Jian Li, Yabiao Wang, Ying Tai, Chengjie Wang, Jilin Li, Feiyue
  Huang, Yili Xia, Wenjiang Pei, and Rongrong Ji.
\newblock Asfd: Automatic and scalable face detector.
\newblock {\em arXiv preprint arXiv:2003.11228}, 2020.

\bibitem{zhang2019accurate}
Faen Zhang, Xinyu Fan, Guo Ai, Jianfei Song, Yongqiang Qin, and Jiahong Wu.
\newblock Accurate face detection for high performance.
\newblock {\em arXiv preprint arXiv:1905.01585}, 2019.

\bibitem{zhang2016joint}
Kaipeng Zhang, Zhanpeng Zhang, Zhifeng Li, and Yu Qiao.
\newblock Joint face detection and alignment using multitask cascaded
  convolutional networks.
\newblock {\em IEEE Signal Processing Letters}, 23(10):1499--1503, 2016.

\bibitem{mtcnn}
Kaipeng Zhang, Zhanpeng Zhang, Zhifeng Li, and Yu Qiao.
\newblock Joint face detection and alignment using multitask cascaded
  convolutional networks.
\newblock {\em IEEE Signal Processing Letters}, 23(10):1499--1503, 2016.

\bibitem{zhang2017detecting}
Kaipeng Zhang, Zhanpeng Zhang, Hao Wang, Zhifeng Li, Yu Qiao, and Wei Liu.
\newblock Detecting faces using inside cascaded contextual cnn.
\newblock In {\em Proceedings of the IEEE International Conference on Computer
  Vision}, pages 3171--3179, 2017.

\bibitem{refineface}
Shifeng Zhang, Cheng Chi, Zhen Lei, and Stan~Z Li.
\newblock Refineface: Refinement neural network for high performance face
  detection.
\newblock {\em arXiv preprint arXiv:1909.04376}, 2019.

\bibitem{zhang2019faceboxes}
Shifeng Zhang, Xiaobo Wang, Zhen Lei, and Stan~Z Li.
\newblock Faceboxes: A cpu real-time and accurate unconstrained face detector.
\newblock {\em Neurocomputing}, 364:297--309, 2019.

\bibitem{faceboxes_improved}
Shifeng Zhang, Xiaobo Wang, Zhen Lei, and Stan~Z Li.
\newblock Faceboxes: A cpu real-time and accurate unconstrained face detector.
\newblock {\em Neurocomputing}, 364:297--309, 2019.

\bibitem{s3fd}
Shifeng Zhang, Xiangyu Zhu, Zhen Lei, Hailin Shi, Xiaobo Wang, and Stan~Z Li.
\newblock S3fd: Single shot scale-invariant face detector.
\newblock In {\em Proceedings of the IEEE International Conference on Computer
  Vision}, pages 192--201, 2017.

\bibitem{zhang2018shufflenet(sufflenet)}
Xiangyu Zhang, Xinyu Zhou, Mengxiao Lin, and Jian Sun.
\newblock Shufflenet: An extremely efficient convolutional neural network for
  mobile devices.
\newblock In {\em Proceedings of the IEEE conference on computer vision and
  pattern recognition}, pages 6848--6856, 2018.

\bibitem{zhu2020progressface}
Jiashu Zhu, Dong Li, Tiantian Han, Lu Tian, and Yi Shan.
\newblock Progressface: Scale-aware progressive learning for face detection.
\newblock In {\em European Conference on Computer Vision}, pages 344--360.
  Springer, 2020.

\bibitem{zhu2020tinaface}
Yanjia Zhu, Hongxiang Cai, Shuhan Zhang, Chenhao Wang, and Yichao Xiong.
\newblock Tinaface: Strong but simple baseline for face detection.
\newblock {\em arXiv preprint arXiv:2011.13183}, 2020.

\end{thebibliography}
}

\end{document}


\title{EResFD: Rediscovery of the Effectiveness of Standard Convolution for Lightweight Face Detection \\(Supplementary Material)}

\maketitle

\section{Implementaion details}
\subsection{Anchor Box Settings}  As shown in Table \ref{tab:anchor_setting}, anchor boxes whose sizes are ranged from 16 to 512 are assigned from detection layers from D1 to D6, respectively, following \cite{s3fd}. The ratio of width to height for each anchor box is set to 1:1.25. 

\begin{table}[h]
    \small
    \centering
    \caption{Anchor box settings: the stride size within the network, anchor box size assigned to 6 detection layers in Figure 2 of the manuscript.}
    \vspace{4mm}
    \label{tab:anchor_setting}
    \resizebox{0.65\linewidth}{!}{
    \begin{tabular}{ccc}
    \toprule
    Detection Layer & Stride & Anchor Size \\
    \midrule
    D1 & 4 & 16 \\
    \midrule
    D2 & 8 & 32 \\
    \midrule
    D3 & 16 & 64 \\ 
    \midrule
    D4 & 32 & 128 \\ 
    \midrule
    D5 & 64 & 256 \\ 
    \midrule
    D6 & 128 & 512 \\ 
    \bottomrule
    
    \end{tabular}
    }
\end{table}

\subsection{Training}
During training\footnote{We used source code from \url{https://github.com/yxlijun/S3FD.pytorch}}, we adopt data pre-processing steps, augmentation strategies, and loss functions used in \cite{s3fd}. Specifically, we used color distortion by changing the hue, saturation, and value (brightness) of an image. Also, for generating various scales of faces, zoom-in and out operations are applied to the image along with its face bounding boxes. Consequently, the resultant image is resized to 640 $\times$ 640 after horizontal flipping. Also, max-out background label~\cite{s3fd} is applied on D1 detection head for reducing false positives with regard to small faces. We also employed multi-task loss, where both classification and regression loss are normalized by the number of positive anchors. We set the balancing parameter between classification and regression loss as 1:1.

For optimization hyper-parameters, we used ADAM~\cite{kingma2014adam} optimizer with initial learning rate 0.001, weight decay 5e-4, and batch size 32. The maximum number of iterations is 330k and the learning rate is decayed at [250k, 300k, 320k] with the decaying factor of 0.1.


\section{Stem Modification}
\begin{table}[t]
    \small
    \centering
    \caption{Latency and detection accuracy according to the stem design. We only changed the stem layer of EResFD with that of ResNet and our proposed stem design of EResNet.}
    \vspace{4mm}
    \smallskip
    \noindent

    \resizebox{1.0\linewidth}{!}{
    \begin{tabular}{c||c|c}
    \hline
    Stem & ResNet & EResNet \\ \hline \hline
    Stem FLOPs & $180.6$ M & $11.5$ M \\ \hline
    Stem Latency (Ratio) & 24.1ms (42\%) & 4.5ms (12\%) \\ \hline
    Overall Latency & 56.8ms & 37.7ms \\ \hline
    Easy mAP (\%) & 87.5 & 89.0 \\ \hline
    Medium mAP (\%) & 86.3 & 87.9 \\ \hline
    Hard mAP (\%) & 77.6 & 80.4 \\ \hline
    \end{tabular}
    }
    \label{table:stemlayer_modification_performance}
\end{table} 
In Table \ref{table:stemlayer_modification_performance}, we compare detection performance depending on the design of the stem layer. It is worth notable that our proposed stem layer requires much more reduced computational costs with significantly improved detection accuracy scores compared to that of ResNet.

\section{SepFPN}
\begin{figure}[t]
    \centering
    \begin{subfigure}[t]{0.5\linewidth}
        \includegraphics[width=1.1\linewidth]{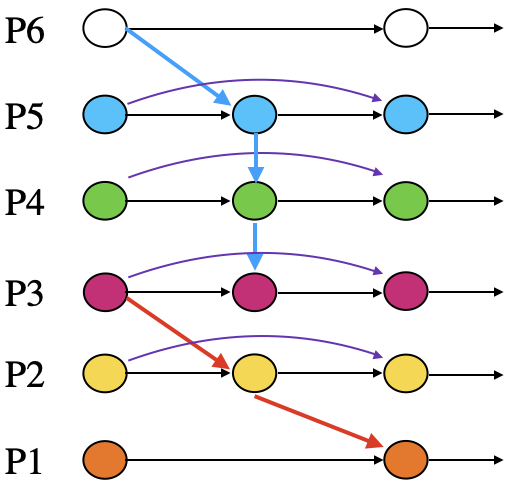}
        \centering
    \subcaption{Separation position: P3}
    \vspace{3mm}
    \end{subfigure}
    \begin{subfigure}[t]{0.5\linewidth}
        \includegraphics[width=1.1\linewidth]{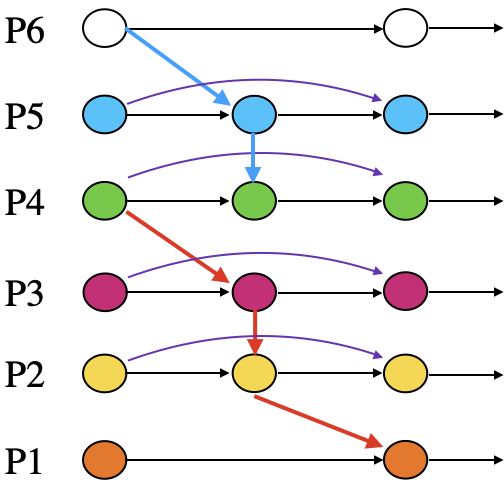}
        \centering
    \subcaption{Separation position: P4}
    \end{subfigure}
    \caption{Architectures of SepFPN with various separation positions (P3, P4).}
    \label{fig:sepfpn_p3_p4}
\end{figure}

 In Figure \ref{fig:sepfpn_p3_p4}, we visualize the architecture of variations for our proposed SepFPN, where its separation position is varied from P3 to P4. 

\newpage

{\small
\bibliographystyle{ieee_fullname}
\bibliography{egbib}
}